%% file: main.tex
\documentclass[10pt,twocolumn,letterpaper]{article}

\usepackage{cvpr}              

\usepackage{kotex}
\usepackage{inconsolata}
\usepackage{pifont}
\AtBeginDocument{\fontdimen2\font=0.9\fontdimen2\font} 
\usepackage{array}
\usepackage{multirow}
\usepackage{adjustbox}
\usepackage{etoolbox}
\usepackage{minitoc}
\usepackage{tocvsec2}
\usepackage{xspace}
\usepackage{appendix}
\usepackage[defaultcolor=magenta]{changes}
\newcommand{\dataset}{MIMIC-ILS\xspace}
\newcommand{\model}{ROSALIA\xspace}
\usepackage{amsmath}
\usepackage{microtype}
\usepackage{chapterbib}
\usepackage{lineno}
\newcommand\blfootnote[1]{%
  \begingroup
  \renewcommand\thefootnote{}\footnote{#1}%
  \addtocounter{footnote}{-1}%
  \endgroup
}

\usepackage[bottom]{footmisc}
\doparttoc

\definecolor{cvprblue}{rgb}{0.21,0.49,0.74}
\usepackage[pagebackref,breaklinks,colorlinks,allcolors=cvprblue]{hyperref}
\usepackage[linesnumbered,ruled,vlined]{algorithm2e}

\def\paperID{} 
\def\confName{CVPR}
\def\confYear{2026}

\title{Instruction-Guided Lesion Segmentation for Chest X-rays \\with Automatically Generated Large-Scale Dataset}

\author{
Geon Choi\textsuperscript{1,\dag} \enspace  Hangyul Yoon\textsuperscript{1,\dag} \enspace Hyunju Shin\textsuperscript{2} \enspace Hyunki Park\textsuperscript{2} \enspace Sang Hoon Seo\textsuperscript{2} \vspace{0.05in} \\ Eunho Yang\textsuperscript{1,3} \enspace Edward Choi\textsuperscript{1,*} \vspace{0.05in}\\
\textsuperscript{1}KAIST\vspace{0.02in} \quad \textsuperscript{2}Samsung Medical Center \quad \textsuperscript{3}AITRICS} 

\begin{document}

\twocolumn[{%
\renewcommand\twocolumn[1][]{#1}%
\maketitle
\vspace{-1.31cm}
\begin{center}
\includegraphics[width=1.0\textwidth]{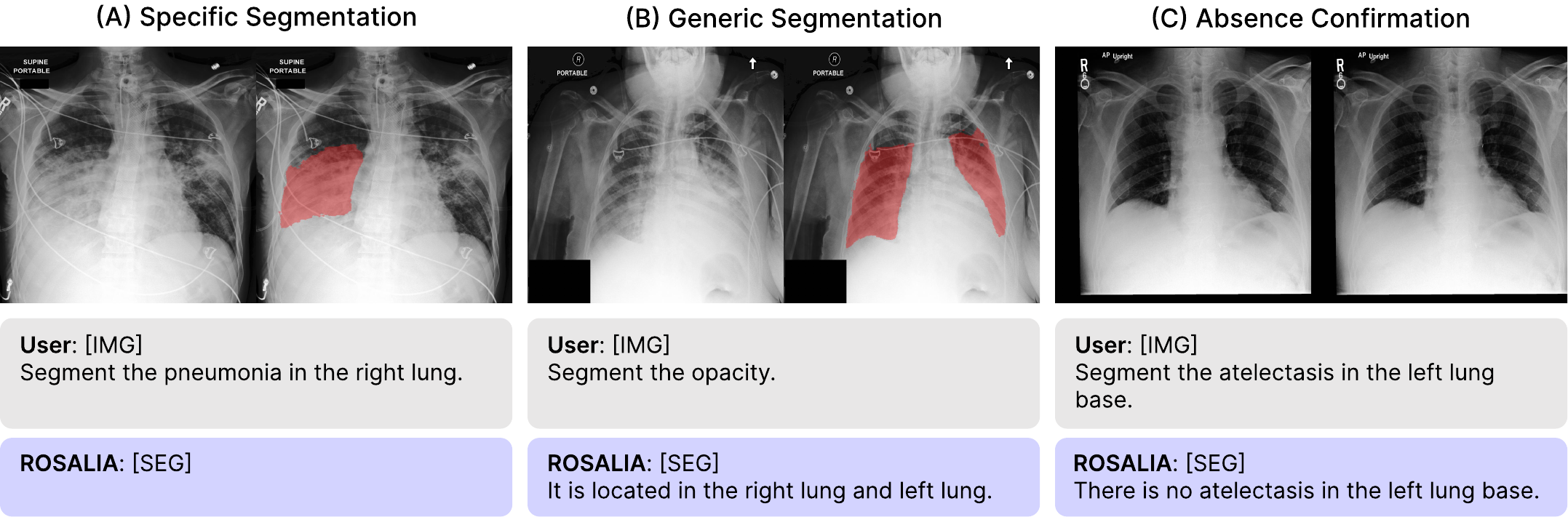}
\vspace{-0.68cm}
    \captionof{figure}{Examples of the instruction-guided CXR lesion segmentation task. Given text instructions for various lesion types and locations of interest, ROSALIA, a VLM trained on our MIMIC-ILS dataset, can: (A) segment lesions in a specified location, (B) segment lesions globally, and (C) detect empty-target cases. As can be seen in (A), ROSALIA correctly ignores the unrequested lesion in the left lung.}
\label{fig:language_guided_lesion_segmentation}
\vspace{-0.12cm}
\end{center}
}]

\begin{abstract}
The applicability of current lesion segmentation models for chest X-rays (CXRs) has been limited both by a small number of target labels and the reliance on complex, expert-level text inputs, creating a barrier to practical use. To address these limitations, we introduce instruction-guided lesion segmentation (ILS), a medical-domain adaptation of referring image segmentation (RIS) designed to segment diverse lesion types based on simple, user-friendly instructions. Under this task, we construct MIMIC-ILS, the first large-scale instruction-answer dataset for CXR lesion segmentation, using our fully automated multimodal pipeline that generates annotations from CXR images and their corresponding reports.
MIMIC-ILS contains 1.1M instruction-answer pairs derived from 192K images and 91K unique segmentation masks, covering seven major lesion types. To empirically demonstrate its utility, we present ROSALIA, a LISA model fine-tuned on the MIMIC-ILS dataset.
ROSALIA can segment diverse lesions and provide textual explanations in response to user instructions. 
The model achieves high accuracy in our newly proposed task, highlighting the effectiveness of our pipeline and the value of MIMIC-ILS as a foundational resource for pixel-level CXR lesion grounding. The dataset and model are available at \href{https://github.com/checkoneee/ROSALIA}{https://github.com/checkoneee/ROSALIA}.

\end{abstract}
\newpage

\input{sec/1_intro}
\input{sec/2_related_work}

\input{sec/3_method}

\input{sec/4_experiments}
\input{sec/5_conclusion}

{
    \small
    \bibliographystyle{ieeenat_fullname}
    \bibliography{main}
}

\input{sec/X_suppl}

\end{document}

%% file: sec/1_intro.tex
\section{Introduction}
\label{sec:intro}
\blfootnote{\textsuperscript{\dag}Equal Contribution \; \textsuperscript{*}Correspondence to}

\vspace{-0.45cm}
Medical imaging is an essential technique in modern medicine, enabling accurate diagnosis and appropriate treatment.
Among various imaging modalities, chest X-ray (CXR) is one of the most common examinations due to its high accessibility and rapid acquisition~\cite{broder2011imaging}.
Radiologists reach a diagnosis by integrating visual evidence from CXRs with their clinical knowledge, and describe these findings in a text format known as a \textit{radiology report}.
A key step in this diagnostic process is identifying the precise location and boundary of a \textit{lesion}—an abnormal region with pathological changes~\cite{coronado2004nci}.
This task is labor-intensive and demands substantial clinical expertise and analytical precision.

\input{table/dataset}

\vspace{-0.03cm}
To alleviate physicians’ workload in localizing pathological regions, there is a growing demand for automated lesion segmentation models in CXRs. 
Recently, vision–language models (VLMs) equipped with segmentation modules~\cite{lai2024lisa, ren2024pixellm, lan2024text4seg} have emerged as a promising solution for referring image segmentation (RIS), as they can interpret diverse user-specific needs expressed through natural language instructions.
However, despite the success of such VLMs in general-domain RIS, their application to CXRs remains limited. 
Although prior studies~\cite{li2023lvit, huang2024cross} have explored CXR lesion segmentation using text prompts, they are limited to a single lesion type (\eg, COVID-19) and moreover require long, detailed expert-level medical descriptions based on tailored CXR review (\eg, ``Bilateral pulmonary infection, two infected areas, upper right lung and upper left lung.'') as input.
Such constraints make them impractical not only for physicians who aim to segment diverse lesion types across various anatomical subregions before closely reviewing the image themselves, but especially for non-experts who can hardly interpret CXR images at all.

To address these limitations, we propose a more user-friendly task, namely \textit{instruction-guided lesion segmentation} (ILS).
In this task, the model is required to process diverse user instructions, ranging from prompts that specify the lesion type and target location, to requests that look for abnormalities globally. If the requested lesion is not present, the model should reliably report its absence. Additionally, the model should be able to provide textual descriptions regarding a lesion's location or type, even if not explicitly prompted by the user. However, a dataset to support such a versatile task has been unavailable, as constructing a suitable dataset for training and evaluation poses significant challenges—most notably the need for expert-curated mask annotations. Moreover, accurately pairing these masks with precise textual instructions in terms of anatomical locations and specific lesion types remains a highly complex task.

In this work, we introduce the first fully automated pipeline for constructing a large-scale ILS dataset for CXRs. The central challenge is: ``How can we derive lesion masks and corresponding instruction-answer text pairs from raw images that contain no explicit annotations?''
To address this, we leverage radiology reports as a key source of information for each image.
Using paired image–report data, our two-stage pipeline integrates pre-trained vision models and large language models (LLMs) to extract high-confidence anomalous regions and structured textual information.
By exploiting the consistency between these heterogeneous modalities, we generate high-quality lesion masks and diverse instruction–answer pairs. Applying our novel framework to MIMIC-CXR~\cite{johnson2019mimic, johnson2024mimic}—a large, publicly available CXR–report dataset—we constructed \textbf{\dataset}, a large-scale dataset consisting of 1.1M samples derived from 192K images and 91K lesion masks (Table~\ref{tab:dataset_summary}).

Although several datasets~\cite{nguyen2022vindr, de2025padchest, boecking2022making, liu2020rethinking, siim-acr-pneumothorax-segmentation, degerli2022osegnet, danilov2022indirect} have tried to introduce spatial annotations in the CXR domain, they are unsuitable for direct use in our ILS task (Table~\ref{tab:dataset_summary}). Most provide only coarse bounding-box localization or single lesion type masks that are limited in scale due to reliance on expert annotations. Moreover, they also lack explicit links between mask annotations and textual instructions.
\dataset bridges these gaps by offering large-scale instruction–answer pairs, each paired with an auto-labeled segmentation mask and a detailed lesion profile. Despite being constructed entirely without human intervention, expert evaluations report a high acceptance rate of over 95\% for this dataset.

Leveraging \dataset, we fine-tune LISA~\cite{lai2024lisa} to develop \textbf{\model} (\textbf{R}adi\textbf{O}logy \textbf{S}egmentation \textbf{A}ssistant trained on a \textbf{L}esion-grounded \textbf{I}nstruction-\textbf{A}nswer dataset), the first VLM designed for ILS task in CXRs.
Given simple and concise user instructions, \model generates segmentation masks and textual descriptions (Fig.~\ref{fig:language_guided_lesion_segmentation}). It supports a wide range of tasks, including specific segmentation (\eg, ``Segment the pneumonia in the right lung.''), generic segmentation (\eg, ``Segment the opacity.''), and absence confirmation (\eg, ``There is no atelectasis in the left lung base.''). This flexibility enables \model to effectively address diverse user needs, delivering tailored outputs for each request. 

In summary, our contributions are threefold:
\begin{itemize}
\item We introduce a novel automated pipeline that generates lesion masks and corresponding instructions directly from CXRs without any human intervention. Using only image–report pairs, our method produces a large-scale dataset without requiring explicit manual processing.

\item Applying our framework to MIMIC-CXR, we construct \dataset, the first dataset for instruction-guided lesion segmentation (ILS) in CXRs. The resulting dataset is further validated by medical experts, confirming its high quality and the reliability of the construction process.

\item To validate the utility of MIMIC-ILS, we introduce \model, the first VLM designed for ILS in CXRs. Trained on our million-scale dataset, \model interprets user instructions across diverse lesion types and locations, producing accurate lesion masks and descriptive outputs. These results demonstrate the value of our dataset and model in advancing fine-grained lesion grounding for CXR analysis.

\end{itemize}
\vspace{-0.1cm}

%% file: table/dataset.tex
\begin{table*}[t]
\centering
\caption{Existing CXR datasets with spatial annotations for pathologic lesions.}
\vspace{-0.25cm}
\label{tab:dataset_summary}
\footnotesize
\resizebox{0.99\linewidth}{!}{%
\begin{tabular}{l c c c c c c}
\toprule
\multirow{2}{*}{\textbf{Dataset}} & 
\multirow{2}{*}{\textbf{\# Images}} & 
\multicolumn{4}{c}{\textbf{Spatial Annotation}} & 
\multirow{2}{*}{\textbf{Instruction-Answer Pair}} \\ 
\cmidrule(lr){3-6}
& & \textbf{\# Annotations} & \textbf{Type} & \textbf{Multi-Lesion} & \textbf{Method} & \\
\midrule
VinDr-CXR~\cite{nguyen2022vindr} & 15K & 9K & Bounding Box & \textcolor{Green}{\ding{51}} & Manual & \textcolor{red}{\ding{55}} \\
Padchest-GR~\cite{de2025padchest} & 4.6K & 7.7K & Bounding Box & \textcolor{Green}{\ding{51}} & Manual & \textcolor{red}{\ding{55}} \\
MS-CXR~\cite{boecking2022making} & 1K & 1.2K & Bounding Box & \textcolor{Green}{\ding{51}} & Manual & \textcolor{red}{\ding{55}} \\
TBX-11K~\cite{liu2020rethinking} & 12K & 1.2K & Bounding Box & \textcolor{red}{\ding{55}} & Manual & \textcolor{red}{\ding{55}} \\
SIIM-ACR~\cite{siim-acr-pneumothorax-segmentation} & 13K & 2.7K & Segmentation Mask & \textcolor{red}{\ding{55}} & Manual & \textcolor{red}{\ding{55}} \\
QaTa-COV19~\cite{degerli2022osegnet} & 121K & 9.3K & Segmentation Mask & \textcolor{red}{\ding{55}} & Semi-Automated & \textcolor{red}{\ding{55}} \\
Danilov et al.~\cite{danilov2022indirect} & 1.4K & 0.6K & Segmentation Mask & \textcolor{Green}{\ding{51}} & Manual & \textcolor{red}{\ding{55}} \\
\midrule
\textbf{MIMIC-ILS (Ours)} & 192K & 91K & Segmentation Mask & \textcolor{Green}{\ding{51}} & Fully-Automated & \textcolor{Green}{\ding{51}} \\
\bottomrule
\end{tabular}
}
\vspace{-0.5cm}
\end{table*}

%% file: sec/2_related_work.tex
\section{Related Work}
\label{sec:related_work}

\begin{figure*}[t]
  \centering
  \includegraphics[width=0.98\textwidth]{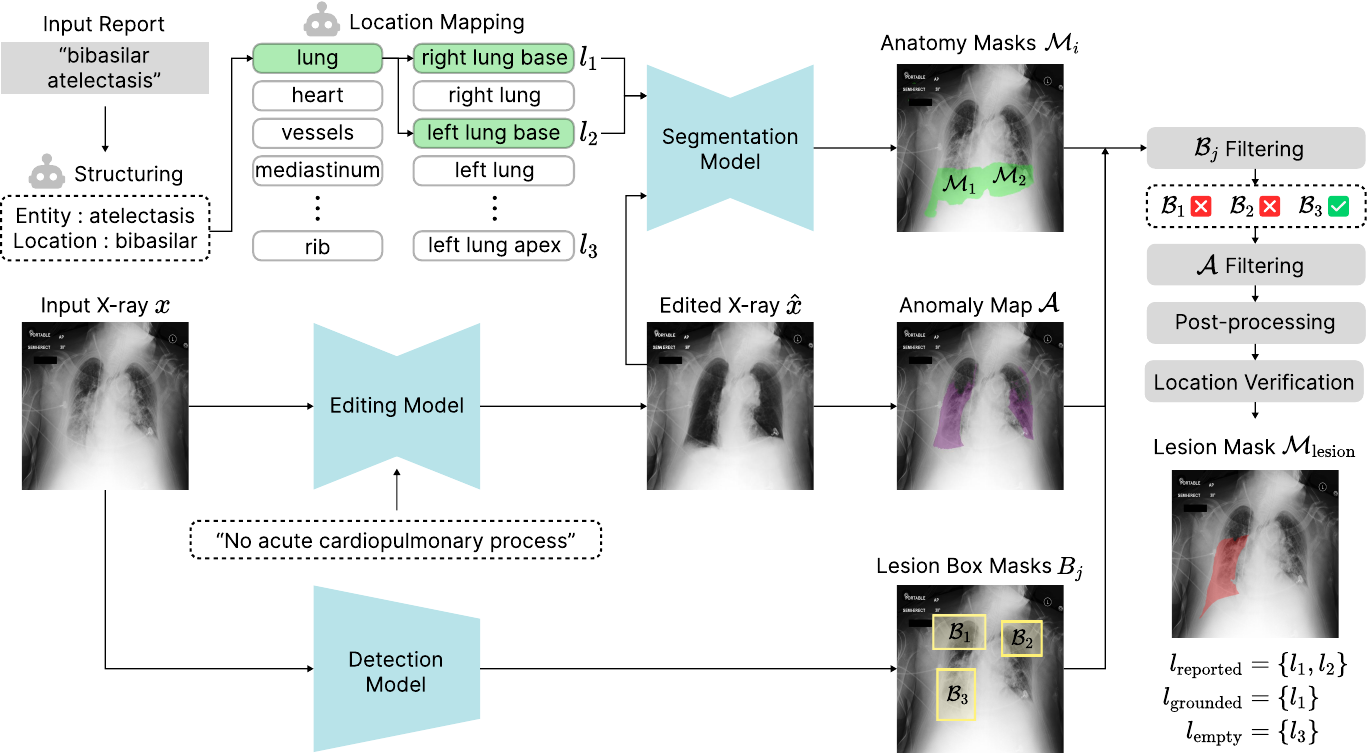}
  \vspace{-0.25cm}
  \caption{An overview of grounded lesion mask generation. \textbf{(Top-left)} Textual information is extracted from the radiology report during the report structuring and location mapping. \textbf{(Bottom-left and Center)} Pretrained vision models are also employed to produce spatial information. \textbf{(Right)} Finally, a lesion mask is generated by integrating this information. The verification step then confirms the grounded location ($l_1$), identifies the empty location ($l_3$) for negative sample generation, and discards the reported-but-ungrounded location ($l_2$).}
  \label{fig:report_grounding}
  \vspace{-0.65cm}
\end{figure*}

\vspace{-0.1cm}
\subsection{Lesion Segmentation and Datasets} 
\vspace{-0.05cm}
Lesion segmentation aims to generate masks corresponding to abnormal regions in medical images. Typically, models are trained on datasets where radiologists have directly annotated lesion masks. For CT and MRI, several studies~\cite{jiang2024zept, jiang2024unleashing} have utilized public datasets that provide diverse tumor masks~\cite{bilic2023liver, heller2020international, antonelli2022medical}. In contrast, such pixel-level annotations are scarce in the CXR domain. While some datasets provide only bounding boxes~\cite{nguyen2022vindr, de2025padchest}, those that offer segmentation masks usually focus on a single lesion type~\cite{liu2020rethinking, degerli2022osegnet, siim-acr-pneumothorax-segmentation}.
Consequently, existing models trained on these datasets are limited in their effective segmentation range for CXRs~\cite{zhao2024biomedparse}.
Our work directly addresses this gap by constructing a comprehensive, multi-type lesion segmentation dataset for CXRs.

\vspace{-0.1cm}
\subsection{Referring Image Segmentation}
\vspace{-0.05cm}
Referring image segmentation (RIS) is the task of segmenting a target specified by text. Early approaches to this task focused on aligning image features with text labels to generate corresponding masks~\cite{li2022language, zhou2022extract, xu2022groupvit, shin2022reco}. More recently, advancements in VLMs have enabled researchers to extend their reasoning capabilities to segmentation~\cite{lai2024lisa, lan2024text4seg, ren2024pixellm}. These models can generate an appropriate mask based on complex instructions that require real-world knowledge, such as ``Segment the object richest in vitamin C in this photo.''

Similar research has emerged in the medical domain, but current approaches remain limited. They usually rely on simple prompts including class labels (\eg, ``a computerized tomography of a tumor'')~\cite{liu2023clip, cheng2025interactive}, which cannot handle sentence-level instructions.
In the CXR domain specifically, recent VLMs have been trained using free-form text that describes the location and number of lesions~\cite{li2023lvit, huang2024cross}.
These approaches, however, expect users to have already reviewed the CXR image, thus providing expert-level descriptions as input.
In contrast, our model allows users to obtain the lesion mask, its presence or absence, and type information even without having to interpret the CXR images first.
\vspace{-0.15cm}

%% file: sec/3_method.tex
\section{Automatic Dataset Construction}
\label{sec:method}
\vspace{-0.1cm}

This section outlines our approach to automatically constructing a large-scale dataset for training a model that generates both lesion segmentation masks and corresponding textual descriptions in response to user instructions. The main challenges in this process are: (1) generating lesion masks directly from raw CXR images without explicit image annotations, (2) aligning appropriate instruction–answer texts with the obtained masks, and (3) ensuring that the entire pipeline operates in a fully automated, human-free manner. 
To address these challenges, our framework first extracts textual and spatial information from image–report pairs and generates lesion masks followed by a verification process (Sec.~\ref{sec:method_report_grounding}).
Using the verified lesion masks, we then construct diverse instruction–answer pairs (Sec.~\ref{sec:method_dynamic_instruction_answer_pair_generation}). 
\vspace{-0.1cm}
\subsection{Grounded Lesion Mask Generation} 
\label{sec:method_report_grounding}
To construct our dataset, we use MIMIC-CXR~\cite{johnson2019mimic, johnson2024mimic}, a large collection of CXR images paired with radiology reports.
Each report is written by a radiologist and provides visual descriptions of the corresponding CXR image.
Based on this dataset, we generate grounded lesion masks through four sequential steps as illustrated in Fig.~\ref{fig:report_grounding}:
(1) Report structuring and location mapping; (2) Spatial information extraction; (3) Lesion mask generation; (4) Location verification. The details of each step are provided in Appendix~\ref{sec:suppl_report_grounding}.

\vspace{5pt}
\noindent\textbf{Report Structuring and Location Mapping.}
The first step employs LLMs to convert radiology reports into a structured form for later steps.
Specifically, we instruct an LLM to transform each sentence describing an abnormal finding into a six-element tuple consisting of the following categories: entity, sentence index, presence, certainty, location, and predicted lesion type. 
The location element is then mapped to one or more anatomical labels to ensure compatibility with the segmentation model used in subsequent processes.
For example, if the second sentence in a given radiology report is ``The lower lung opacity is pneumonia.", its corresponding output is \texttt{(opacity, 2, positive, definitive, [right lung base, left lung base], pneumonia)}.
Here, the term ``lower lung'' in the original report is mapped to ``right lung base'' and ``left lung base''.

\vspace{5pt}
\noindent\textbf{Spatial Information Extraction.} The second step extracts spatial information from CXRs using three distinct models: (1) RadEdit~\cite{perez2024radedit}, a diffusion-based image editing model; (2) CXAS~\cite{seibold2023accurate}, an anatomy segmentation model; and (3) a pretrained YOLO model for CXR lesion detection~\cite{Nguyen2025Localizing}.
These models are used respectively to generate an anomaly map, anatomy masks, and lesion box masks, which serve as visual cues for lesion mask generation in the subsequent steps.

RadEdit takes an input image $x \in \mathbb{R}^{H \times W}$ containing a lesion and the text prompt ``No acute cardiopulmonary process'' and outputs an edited image $\hat{x}$ from which the lesion has been removed. We derive $x_\mathrm{ano} \in [0, 1]^{H \times W}$
as:
\begin{equation*}
x_\mathrm{ano} = \frac{x - \hat{x}}{I_\mathrm{max}},
\label{eq:anomaly_map}
\end{equation*}
where $I_{\text{max}}$ is the maximum possible pixel intensity (\ie, 255 for an 8-bit image).
$x_\mathrm{ano}$ provides morphological information about hyperintense lesions, which are areas that appear brighter than the normal lung field. From $x_\mathrm{ano}$, we define anomaly map $\mathcal{A}$ as:
\begin{equation}
\mathcal{A} = {\{(i, j) \mid (x_\mathrm{ano})_{i,j} \ge \tau_\mathrm{ano} \}},\label{eq:anomaly_set}
\end{equation}
where $(i,j)$ represents a pixel coordinate and $\tau_\mathrm{ano}$ is a threshold for anomaly pixels. 

CXAS produces anatomy masks corresponding to each location element in the previously derived structured report tuples. We denote these masks as coordinate sets $\{\mathcal{M}_i\}_{i=1}^n$, where $n$ is the number of anatomical labels mapped in the previous step. Each $\mathcal{M}_i$ contains the pixel coordinates for a specific anatomy, serving as a spatial approximation of the lesion location mentioned in the radiology report.

In parallel, the pretrained YOLO model is applied to $x$ to detect a diverse range of lesions. It outputs bounding boxes that not only specify the locations of potential lesions but also assign a confidence score to each detection. From these results, we construct a set of lesion box masks, $\{\mathcal{B}_j\}_{j=1}^m$, where $m$ denotes the number of detected boxes. Each $\mathcal{B}_j$ represents the pixel coordinates enclosed by a bounding box, accompanied by a confidence score $\mathit{conf}_{\mathcal{B}_j} \in [0,1]$.

\vspace{-0.1cm}
\vspace{5pt}
\noindent\textbf{Lesion Mask Generation.} With the three visual cues extracted from the previous step, the initial lesion masks can be generated.
Here, the anomaly map $\mathcal{A}$ plays a central role, representing a composite signal of all hyperintense lesions.
We decompose this signal into individual masks and align them with the specific lesions described in the report.
During this process, the anatomy masks $\{\mathcal{M}_i\}_{i=1}^n$, lesion box masks $\{\mathcal{B}_j\}_{j=1}^m$, and the right and left lung masks ($L_r$ and $L_l$) are jointly used to select high-quality mask candidates.

The core of this filtering process, outlined in Algorithm~\ref{alg:lesion_mask_generation}, selectively retains only appropriate candidates from the initially detected lesion box masks, based on four conditions:
($\textit{c}_1$) sufficient overlap with $\{\mathcal{M}_i\}_{i=1}^n$;
($\textit{c}_2$) a high confidence score; 
($\textit{c}_3$) a high internal signal ratio from $\mathcal{A}$ (\ie, the ratio of the intersection area between the box mask and $\mathcal{A}$ to the area of the box mask);
and ($\textit{c}_4$) a sufficient size relative to either $L_r$ or $L_l$.
Conditions $\textit{c}_1$ and $\textit{c}_2$ ensure that the boxes align with the reported locations and are likely to correspond to true lesions.
However, the $\mathcal{A}$ can contain false negatives (\ie, coordinates that belong to actual lesion areas but are missing from $\mathcal{A}$), which may result in excessively small or even empty masks.
To mitigate this issue, conditions $\textit{c}_3$ and $\textit{c}_4$ are used to retain only those boxes that contain strong lesion signals and are large enough to allow meaningful segmentation. 
Once the appropriate lesion box masks are selected, we extract from $\mathcal{A}$ the connected components (\ie, the individual, contiguous `islands' in 2D space) that intersect with these selected masks.
This component then undergoes a post-processing step involving small, noisy mask removal to produce the final, refined lesion mask $\mathcal{M}_\mathrm{lesion}$ (see Appendix~\ref{sec:suppl_post_processing} for further details).

\input{algorithm/mask_generation}

\begin{figure*}[ht]
  \centering
  \includegraphics[width=0.98\textwidth]{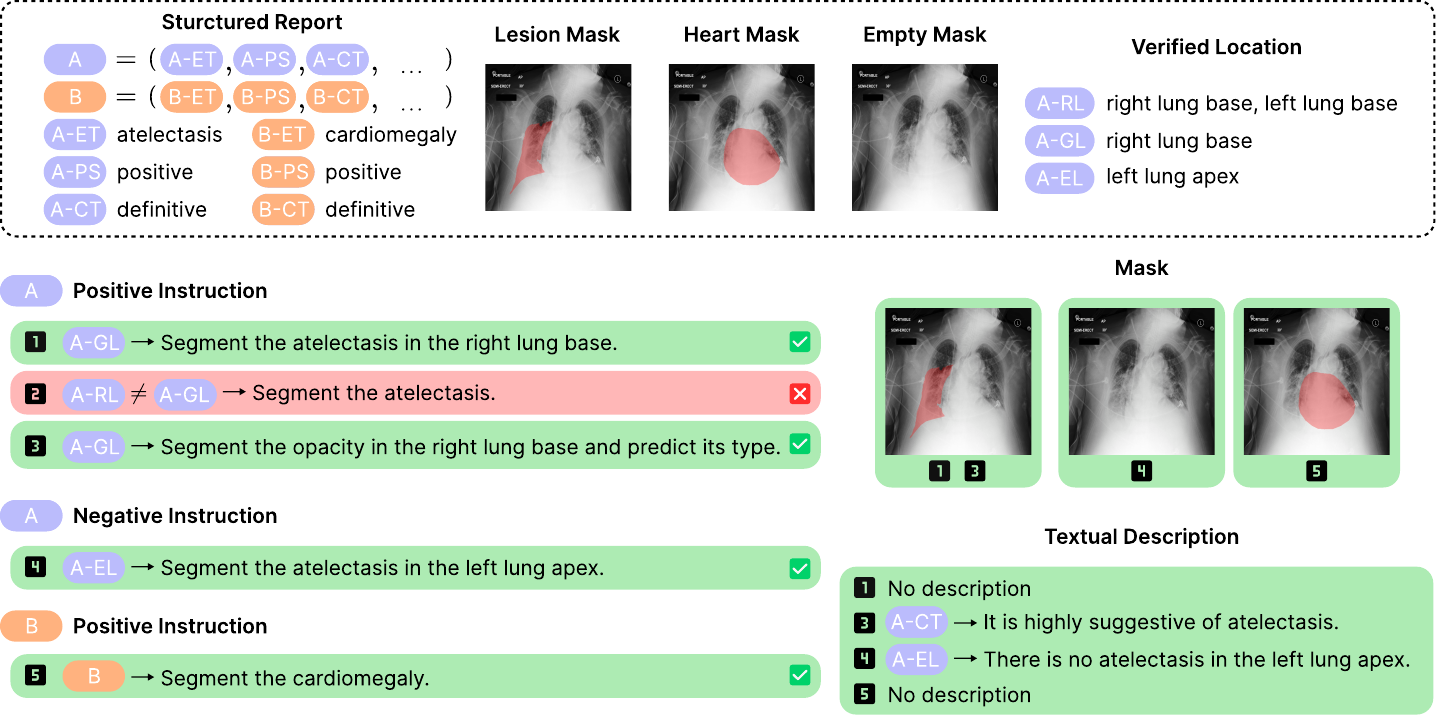}
  \vspace{-0.2cm}
  \caption{Instruction–answer pair generation process using the example report, “Bibasilar atelectasis. Cardiomegaly.”
  We utilize the elements extracted from the previous lesion mask generation
  process (see Fig.~\ref{fig:report_grounding}), indicated by the dashed box.
  Structured tuples (A\&B in the top left) are converted to text instructions and mapped to their corresponding ground-truth masks and textual descriptions.
  Invalid instructions for lesions which lack a corresponding mask are excluded (colored as red), and only valid instructions are retained  (colored as green). (ET: entity, PS: presence, CT: certainty, RL: reported location, GL: grounded location, EL: empty location)}
  \label{fig:instruction_answer_pair}
  \vspace{-0.5cm}
\end{figure*}

\vspace{1pt}
\noindent\textbf{Location Verification.} In the final step, we explicitly verify whether each lesion mask generated by Algorithm~\ref{alg:lesion_mask_generation} has been successfully grounded to the structured report. To assess the grounding status, we define three types of locations: \textit{reported location}, \textit{grounded location} and \textit{empty location}. The \textit{reported location} is a set of anatomical labels extracted from the previous location mapping with LLMs.
Based on this set, the \textit{grounded location} is defined as a subset of the \textit{reported location} that spatially overlaps with a generated lesion mask, confirming successful localization of the reported finding. This location is derived from the anatomy masks $\{\mathcal{M}_i\}_{i=1}^n$ that intersect with the selected lesion box masks during the lesion mask generation. Finally, we introduce an \textit{empty location}, which refers to a lung region with no reported lesions and is used to generate negative samples.

\subsection{Instruction-Answer Pair Generation}
\label{sec:method_dynamic_instruction_answer_pair_generation}
\vspace{-0.15cm}
With the information extracted from the previous process (\ie, grounded lesion mask generation), we build our dataset for seven major lesion types found in CXRs: cardiomegaly, pneumonia, atelectasis, opacity, consolidation, edema, and effusion. 
These lesions are not only the most frequently mentioned in radiology reports, but also clinically significant to be common annotation targets in other medical datasets~\cite{wang2017chestx, johnson2019mimic, boecking2022making, johnson2024mimic}.
For each lesion, we construct positive instruction-answer pairs, which include a ground-truth lesion mask. Negative pairs using an empty mask are also generated to enable the model to confirm the absence of lesions.
An example of this pair generation process is shown in Fig.~\ref{fig:instruction_answer_pair}. 
Please refer to Appendix~\ref{sec:suppl_lesion_types} and~\ref{sec:suppl_instruction_answer_pair_generation} for the lesion descriptions and specific dataset generation process. 

\vspace{3pt}
\noindent\textbf{Instruction Types and Limitations.} We consider three types of segmentation instructions (Table~\ref{tab:template}). A \textit{basic} instruction specifies both the segmentation target and its location.
The location can be a broad region (such as left lung or right lung), one of eight more specific zones (apical, upper, mid, and lower zones for each lung), or a combination of these regions.
In contrast, a \textit{global} instruction specifies only the segmentation target.
A \textit{lesion inference} instruction asks the model to predict the type of lesion represented by an opacity within a given location.
The generation of these instructions is inherently constrained by the grounded lesion mask generation.
For example, a global instruction becomes invalid if the generated mask captures only part of the lesion.
To address this, our framework dynamically produces only those instruction–answer pairs that are valid given the grounding information available for each image.
 
\input{table/template}

\vspace{5pt}
\noindent\textbf{Instruction Generation.} The instruction generation process begins by creating a basic instruction for each grounded lesion. 
Next, we determine whether a global instruction can be generated.
The global instruction is created only when the \textit{grounded location} and the \textit{reported location} are identical. 
Separately, we generate lesion inference instructions by transforming the basic instructions for \textit{pneumonia, atelectasis}, and \textit{edema}, replacing these specific lesion types with \textit{opacity}.
This transformation is motivated by the fact that these findings are all specific types of ``opacity,'' a more fundamental visual concept in medical imaging. Negative samples are generated by (1) selecting lesion types that are not mentioned or explicitly negated in the radiology report; or (2) utilizing \textit{empty locations} to substitute the original location in the basic instruction of a positive sample.

\vspace{2pt}
\noindent\textbf{Answer Generation.} Each answer consists of a lesion mask and a textual description. The answer lesion masks for positive pairs are determined differently depending on whether they are organ-level or localized abnormalities.
For cardiomegaly, we utilize a heart mask as its corresponding lesion mask since this condition is defined by the state of a specific organ~\cite{Gaggion_2022}.
In contrast, localized abnormalities (\eg, pneumonia or effusion) can appear in variable locations, so for these findings, we use the lesion masks generated in Sec.~\ref{sec:method_report_grounding}.
For negative pairs, an empty mask is used.
As for the textual description, it is also provided for both positive and negative samples. Specifically, the answer template for lesion inference incorporates a certainty level.

\vspace{-0.05cm}
\section{MIMIC-ILS Dataset}
\label{sec:method_mimic_ls}

\begin{figure}[h]
  \centering
  \vspace{-0.2cm}
  \includegraphics[width=0.96\linewidth]{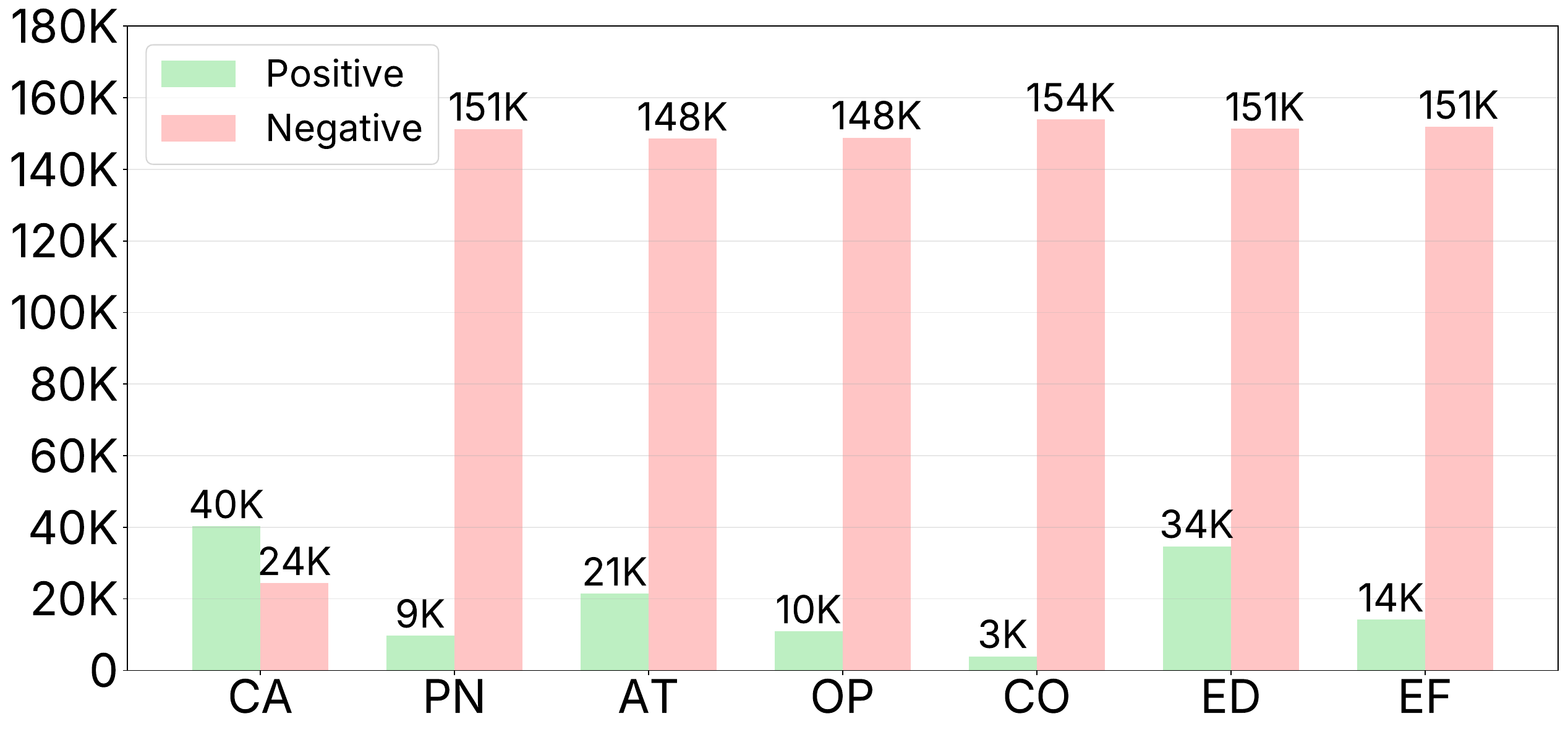}
  \vspace{-0.3cm}
  \caption{Distribution of MIMIC-ILS dataset. The y-axis indicates the number of samples, and the x-axis represents the lesion type. (CA: cardiomegaly, PN: pneumonia, AT: atelectasis, OP: opacity, CO: consolidation, ED: edema, EF: effusion)}
  \label{fig:data_distribution}
  \vspace{-0.45cm}
\end{figure}

Our final dataset, \dataset, consists of 1.1M instruction-answer pairs (135K positive and 930K negative samples) derived from 192K MIMIC-CXR images. 
This final image set is obtained by first filtering out low-quality images (\eg, images with extreme contrast issues), and then excluding any images for which no instruction-answer pairs are generated through our pipeline in Sec.~\ref{sec:method}. The positive samples are generated from 91K unique lesion masks, where each mask can be associated with multiple instruction–answer samples. The resulting dataset covers seven distinct lesion types, and the overall statistics are presented in Fig.~\ref{fig:data_distribution}. Following the official MIMIC-CXR split, the dataset is divided into 1M training samples, 8.2K validation samples, and 12K test samples. Details on quality control and a distribution of \dataset are presented in Appendix~\ref{sec:mimic_quality_control} and~\ref{sec:mimic_ils_dataset}.

\vspace{5pt}
\noindent\textbf{Human Evaluation.} To assess the quality of \dataset, an expert review was conducted by four radiation oncologists specializing in lesion contouring on medical images.
For the test set samples, clinicians classified each case as either acceptable or unacceptable based on mask quality. Positive cases were reviewed by all experts, while negatives were split among them.
Any sample judged unacceptable by at least one expert was excluded from the final test set, and the results are summarized in Table~\ref{tab:human_eval}.
Among the 10.7K mask samples initially reviewed, 96.4\% were rated as acceptable and finally included in the test set. More details on the expert profiles and quality assessment are provided in Appendix~\ref{sec:mimic_ils_dataset}.

\input{table/human_eval}

\section{Model Training}
\label{sec:method_training}
\begin{figure}[h]
  \centering
  \includegraphics[width=0.97\linewidth]{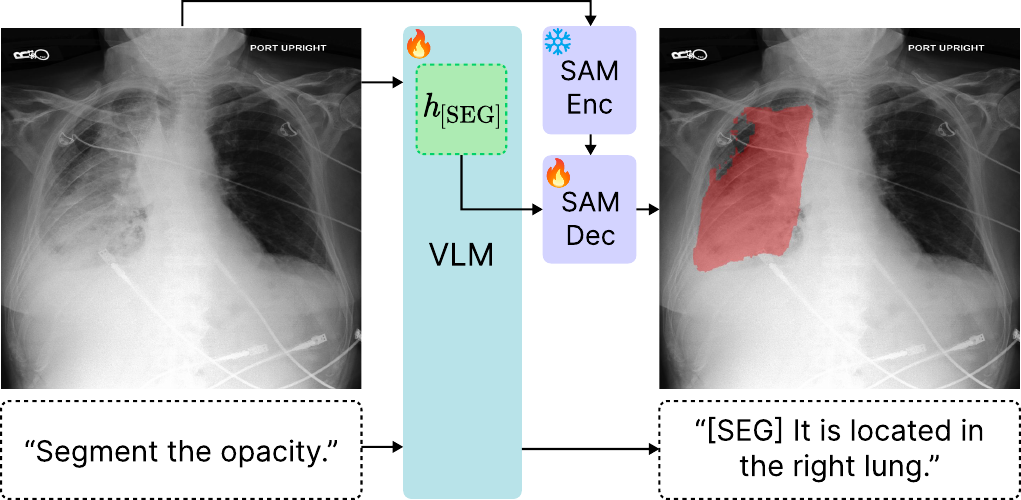}
  \vspace{-0.1cm}
  \caption{Overview of ROSALIA. The architecture integrates a VLM with the SAM. The VLM takes a CXR image and a segmentation instruction as input, generating both a textual description and a special [SEG] token. The hidden embedding of this [SEG] token is then passed to SAM's decoder to produce the final mask.}
  \label{fig:rosalia}
  \vspace{-0.3cm}
\end{figure}

Using \dataset, we train our ILS model, \model. The model adopts the architecture of LISA~\cite{lai2024lisa}, which demonstrated strong zero-shot language-guided segmentation performance in the general domain.
As illustrated in Fig.~\ref{fig:rosalia}, the architecture integrates a VLM backbone with the Segment Anything Model (SAM)~\cite{kirillov2023segment}.
The VLM processes both the image and the input instruction to produce a special token, [SEG], along with its textual description.
This [SEG] token embedding is then passed to SAM together with the input image for mask prediction.
Within SAM, the frozen image encoder extracts embeddings from the image, and the mask decoder integrates these embeddings with the hidden embedding of [SEG] token to generate the final mask.
\vspace{-0.2cm}

The overall loss function $\mathcal{L}$ consists of two components: (1) a language loss and (2) a mask loss. It is formulated as:
\begin{align*} 
\mathcal{L} &= \lambda_{\text{txt}}\mathcal{L}_{\text{txt}} + \mathcal{L}_{\text{mask}}, \\
\mathcal{L}_{\text{mask}} &= \lambda_{\text{bce}}\mathcal{L}_{\text{bce}} + \lambda_{\text{dice}}\mathcal{L}_{\text{dice}}.
\label{eq:loss_term} 
\end{align*}
$\mathcal{L}_{\text{txt}}$ denotes the autoregressive cross-entropy loss for the answer text, and $\mathcal{L}_{\text{mask}}$ represents the segmentation loss computed between the ground-truth mask and the predicted foreground probability map, which combines the binary cross-entropy loss $\mathcal{L}_{\text{bce}}$ and the DICE~\cite{milletari2016v} loss $\mathcal{L}_{\text{dice}}$. The $\lambda_{\text{txt}}$, $\lambda_{\text{bce}}$, and $\lambda_{\text{dice}}$ are coefficients for each loss term.

%% file: algorithm/mask_generation.tex
\setlength{\textfloatsep}{1pt}
\begin{algorithm}[h]
\fontsize{8.6pt}{9.5pt}\selectfont 
\caption{Lesion Mask Generation}
\label{alg:lesion_mask_generation}
\KwIn{Anomaly map $\mathcal{A}$, anatomy masks $\{\mathcal{M}_i\}_{i=1}^n$, lesion box masks $\{\mathcal{B}_j\}_{j=1}^m$ with confidences $\{\mathit{conf}_{\mathcal{B}_j}\}_{j=1}^m$, right lung mask $L_r$, left lung mask $L_l$}
\KwOut{Final lesion mask $\mathcal{M}_\mathrm{lesion}$}
\BlankLine
$\mathcal{M}_\mathrm{lesion} \leftarrow \emptyset$\;
$\mathcal{M}_\mathrm{union} \leftarrow \bigcup_{i=1}^{n} \mathcal{M}_i$\;
\ForEach{$\mathcal{B}_j \in \{\mathcal{B}_j\}_{j=1}^m$}{
    $c_1 \leftarrow \cfrac{|\mathcal{B}_j \cap \mathcal{M}_\mathrm{union}|}{|\mathcal{B}_j \cup \mathcal{M}_\mathrm{union}|} \ge \tau_\mathrm{anatomy}$\;
    $c_2 \leftarrow \mathit{conf}_{\mathcal{B}_j} \ge \tau_\mathrm{conf}$\;
    $c_3 \leftarrow \cfrac{|\mathcal{B}_j\cap \mathcal{A}|}{|\mathcal{B}_j|} \ge \tau_\mathrm{signal}$\;
    $c_4 \leftarrow \left(\cfrac{|\mathcal{B}_j\cap L_r|}{|\mathcal{B}_j \cup L_r|} \ge \tau_\mathrm{size}\right) \lor \left(\cfrac{|\mathcal{B}_j\cap L_l|}{|\mathcal{B}_j \cup L_l|} \ge \tau_\mathrm{size}\right)$\;
    
    \If{$c_1 \land c_2 \land c_3 \land c_4$}{
        $C \leftarrow \text{FindIntersectingComponent}(\mathcal{B}_j, \mathcal{A})$\;
        \If{$C$ is not empty}{
            $\mathcal{M}_\mathrm{new} \leftarrow \text{Refine}(C)$\;
            $\mathcal{M}_\mathrm{lesion} \leftarrow \mathcal{M}_\mathrm{lesion} \cup \mathcal{M}_\mathrm{new}$\;
        }
    }
}
\Return{$\mathcal{M}_\mathrm{lesion}$}
\end{algorithm}

%% file: table/template.tex
\begin{table}[ht]
  \vspace{-0.2cm}
  \caption{Templates for each question type. Each type includes answer templates for both positive and negative cases, with the negative answers positioned in the last row of each cell.}
  \vspace{-0.2cm}
  \label{tab:template}
  \centering
  \footnotesize
  \resizebox{\linewidth}{!}{%
  \begin{tabular}{m{0.21\linewidth}m{0.11\linewidth}p{0.6\linewidth}}
    \toprule
    Type & Role & Template \\
    \midrule
    \multirow{3}{*}{Basic}
      & Instruction & Segment the [Target] in the [Location].\\
      \cmidrule(lr){2-3}
      & \multirow{2}{*}{Answer} & [SEG]\\
      &  & [SEG] There is no [Target] in the [Location].\\
    \midrule
    \multirow{3}{*}{Global}
      & Instruction & Segment the [Target].\\
      \cmidrule(lr){2-3}
      & \multirow{2}{*}{Answer} & [SEG] It is located in the [Location].\\
      &  & [SEG] There is no [Target].\\
    \midrule
    \multirow{4}{*}{\vspace{-0.3cm}Lesion Inference}
      & \multirow{2}{*}{Instruction} & Segment the opacity in the [Location] and predict its type.\\
      \cmidrule(lr){2-3}
      & \multirow{3}{*}{Answer} & [SEG] It is highly suggestive of [Lesion].\\
      &  & [SEG] It possibly reflects [Lesion].\\
      &  & [SEG] There is no opacity in the [Location]. \\
    \bottomrule
  \end{tabular}
  }
\vspace{-0.4cm}
\end{table}

%% file: table/human_eval.tex
\begin{table}[h]
\vspace{-0.2cm}
\caption{Acceptance rate and number of evaluated samples for the human evaluation. Each sample corresponds to a unique combination of lesion mask, target, and location.}
\vspace{-0.2cm}
\normalsize
\centering
\setlength{\tabcolsep}{3pt}
\begin{adjustbox}{max width=0.99\linewidth}
\begin{tabular}{l|cc|cc|cc}
\toprule
\multirow{2}{*}{Expert} & 
\multicolumn{2}{c|}{Total} & 
\multicolumn{2}{c|}{Positive} & 
\multicolumn{2}{c}{Negative} \\
\cmidrule(lr){2-3} \cmidrule(lr){4-5} \cmidrule(lr){6-7}
 & Rate (\%) & \# Samples & Rate (\%) & \# Samples & Rate (\%) & \# Samples \\
\midrule
Expert A & 96.1 & 4,090 & 95.6 & 1,841 & 96.5 & 2,249 \\
Expert B & 97.2 & 4,028 & 96.0 & 1,841 & 98.3 & 2,187 \\
Expert C & 98.7 & 4,041 & 99.8 & 1,841 & 97.8 & 2,200 \\
Expert D & 97.6 & 4,065 & 96.9 & 1,841 & 98.2 & 2,224 \\
\midrule
Overall & 96.4 & 10,701 & 90.1 & 1,841 & 97.7 & 8,860 \\
\bottomrule
\end{tabular}
\end{adjustbox}
\label{tab:human_eval}
\vspace{-0.5cm}
\end{table}

%% file: sec/4_experiments.tex
\section{Experiments}
\label{sec:experiments}

\subsection{Implementation Details}
\label{sec:implementation details}
\noindent\textbf{Training Details.} \model is built on the LISA-7B architecture and is fine-tuned from its original checkpoint~\cite{lai2024lisa}.
Following LISA, we adopted LLaVA~\cite{liu2023visual} as the VLM backbone and employed the largest version of SAM (SAM-H).
LoRA~\cite{hu2022lora} fine-tuning was applied to the VLM with a rank of 128 and an alpha of 256, while the mask decoder was fully fine-tuned.
The epochs and the initial learning rate were set to 15 and 0.0003, respectively, using the AdamW optimizer~\cite{loshchilov2017decoupled}.
The total batch size was 256, and the ratio of positive to negative samples was maintained at 1:1 in each mini-batch.
The loss coefficients $\lambda_{\text{txt}}$, $\lambda_{\text{bce}}$, and $\lambda_{\text{dice}}$ were set to 0.5, 5, and 1, respectively, and the DICE loss was computed only for positive samples. 
Further model training details are described in the Appendix~\ref{sec:suppl_model_training}.

\vspace{5pt}
\noindent\textbf{Baseline Models.} Since we present the first dataset for ILS in CXRs, no existing models have been directly trained on our proposed task. Nonetheless, we evaluated several models from both the general domain (LISA~\cite{lai2024lisa}, Text4Seg~\cite{lan2024text4seg}, PixelLM~\cite{ren2024pixellm}) and the medical domain (BiomedParse~\cite{zhao2024biomedparse}, RecLMIS~\cite{huang2024cross}, IMIS-Net~\cite{cheng2025interactive}), which can take an image and text as input to produce a segmentation output.

\vspace{5pt}
\noindent\textbf{Evaluation Metrics.} We used three metrics to evaluate model performance.
For positive samples, we used Intersection-over-Union (IoU)–based measures: gIoU and cIoU~\cite{lai2024lisa}.
gIoU is the average IoU across samples, while cIoU is the ratio of total intersection to total union across the dataset.
For negative cases, we used empty-target accuracy (N-Acc.), the proportion of samples correctly predicted to have no masks~\cite{xia2024gsva}.

\begin{figure*}[t]
  \centering
  \includegraphics[width=0.99\textwidth]{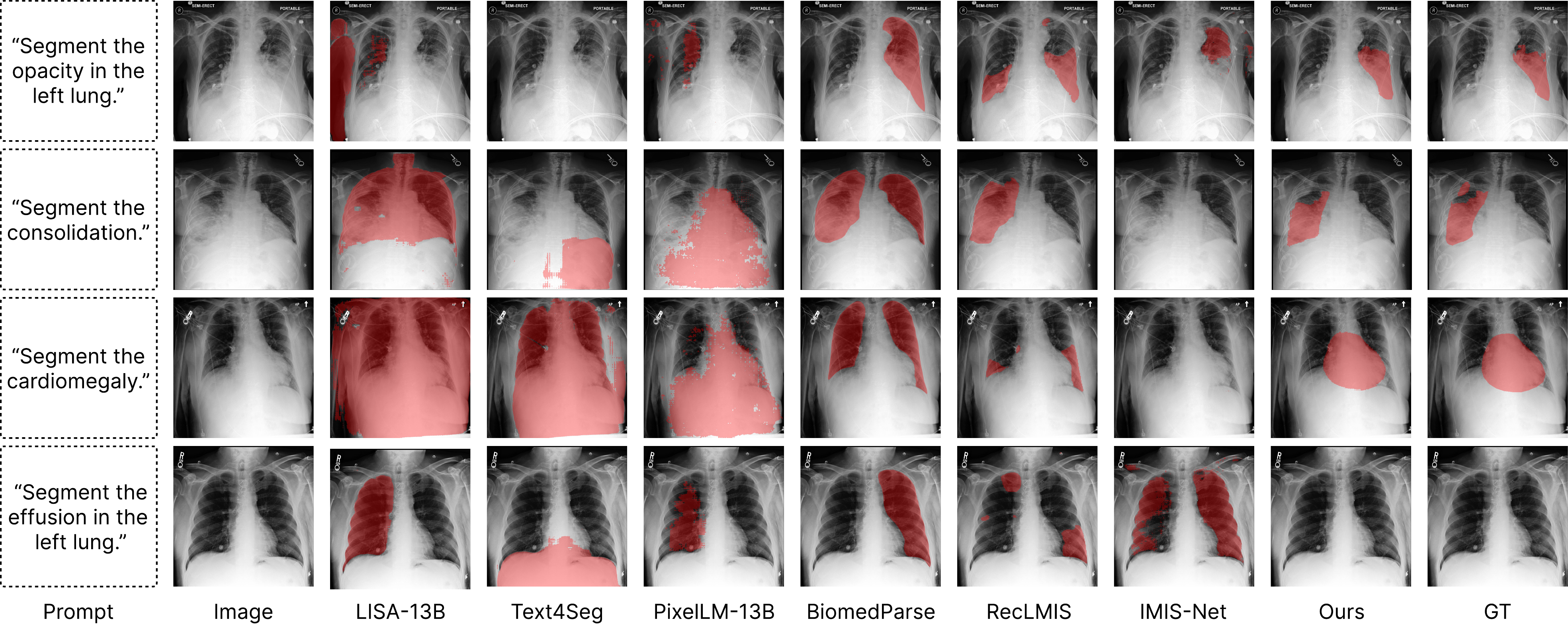}
  \vspace{-0.2cm}
  \caption{Visualized inference results of ROSALIA and baseline models. The first three rows show results for positive cases, while the last row presents results for negative cases with an empty target mask. Additional examples are demonstrated in Appendix~\ref{sec:additional result}.}
  \label{fig:grid}
  \vspace{-0.4cm}
\end{figure*}

\subsection{Main Results}
Table~\ref{tab:main result} presents the results of the baselines and our proposed model on the \dataset test set.
While existing VLM-based segmentation models from both the general and medical domains struggle with the ILS task, \model (LISA-7B fine-tuned on MIMIC-ILS) achieves notably high performance.
In particular, not only do these baselines yield low IoU scores on positive cases, but they also frequently fail on empty-target cases, where the N-Acc. rate is nearly zero in most instances. These results highlight the need for a dedicated dataset to effectively address the ILS task in CXRs.
Furthermore, the strong results of \model on the physician-verified test set demonstrate that the training set of \dataset serves as a high-quality resource—even without manual expert filtering.

\input{table/main}

Table~\ref{tab:main result by lesion} presents the performance of ROSALIA across different lesion types.
The overall gIoU exceeds 0.7, indicating that more than 80\% of the regions overlap between the predicted and ground-truth masks when the two are of similar size.
Even for the lesion type with the lowest gIoU, the score remains above 0.55, suggesting over 70\% regional overlap under similar mask sizes between the ground truth and predictions. 

\input{table/main_lesion}
\vspace{0.1cm}
We also evaluate the accuracy of text responses across different question types, as shown in Table~\ref{tab:text_result}.
A response is considered correct only when both the template and all variables for each question type (denoted by square brackets in Table~\ref{tab:template}) exactly match the structured ground-truth information.
Despite this strict criterion, \model achieves high accuracy across most question types (see Appendix~\ref{sec:additional result} for text accuracy of each lesion type).

\input{table/text_accuracy_class}

\subsection{Qualitative Results}
\label{sec:qualitative_results}

\begin{figure}[h]
  \vspace{-0cm}
  \centering
  \includegraphics[width=0.99\linewidth]{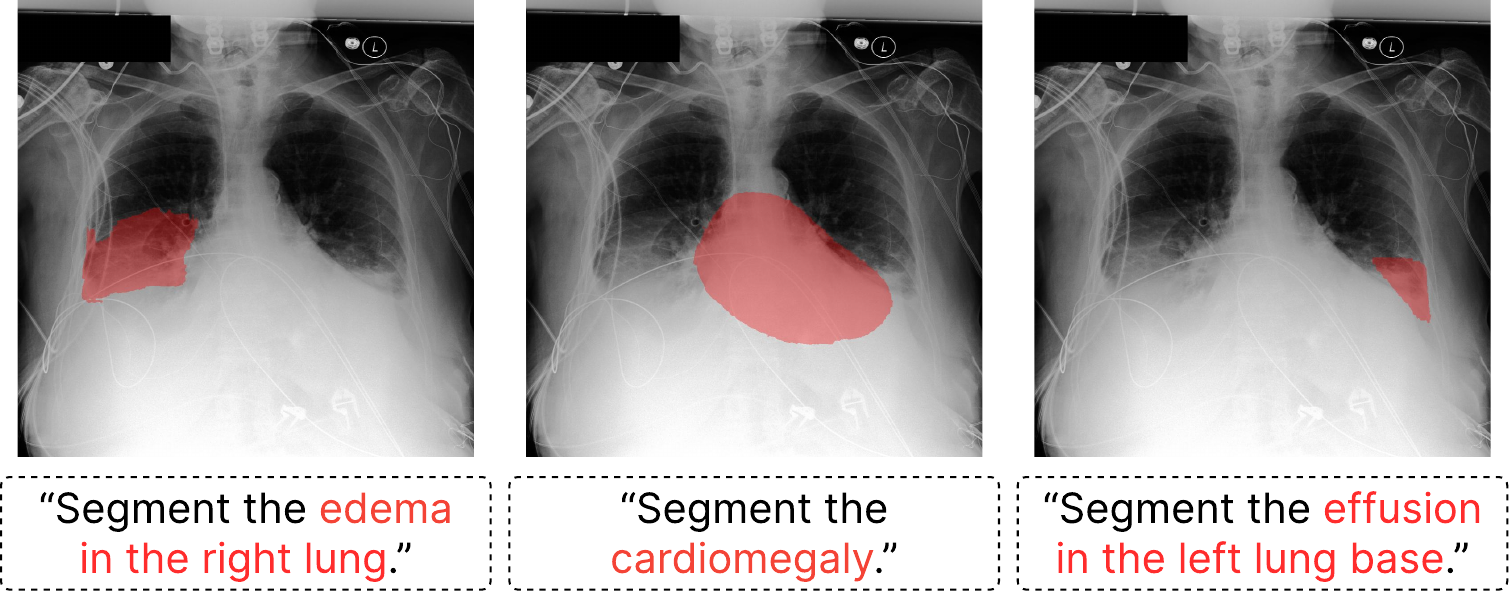}
  \vspace{-0.2cm}
  \caption{Examples of outputs from different instructions applied to the same image. Among the multiple lesions present, ROSALIA can selectively segment only the lesion and location of interest.}
  \label{fig:multi_seg}
  \vspace{-0.4cm}
\end{figure}

Fig.~\ref{fig:grid} presents qualitative examples from each model for the ILS task.
The baseline models largely fail, either producing entirely incorrect masks or segmenting the whole anatomical regions (\eg, the left or right lung).
In contrast, \model accurately segments only the lesion specified in the instruction within the designated region and correctly identifies empty-target cases. Additionally, Fig.~\ref{fig:multi_seg} demonstrates the outputs produced from diverse instructions applied to the same input image.
Although multiple lesions coexist in the image, ROSALIA accurately interprets each instruction and generates results tailored to the user’s specific request. This highlights the model's ability to handle diverse lesion types and locations of interest.

%% file: table/main.tex
\begin{table}[h]
\vspace{-0.2cm}
\caption{Segmentation results (\%) on the MIMIC-ILS test set. ``N-Acc." denotes the accuracy of correctly predicting empty targets. $\P$ indicates medical domain baselines. The best and second-best results are marked in \textbf{bold} and \underline{underline}, respectively.}
\vspace{-0.2cm}
\small
\centering
\begin{adjustbox}{max width=0.73\linewidth}
\begin{tabular}{c|ccc}
\toprule
Model & gIoU & cIoU & N-Acc. \\
\midrule
LISA-7B~\cite{lai2024lisa} & 8.3 & 12.8 & 0.7 \\
LISA-13B~\cite{lai2024lisa} & 8.9 & 12.2 & 0.0 \\
Text4Seg~\cite{lan2024text4seg} & 6.1 & 10.3 & 20.6 \\
PixelLM-7B~\cite{ren2024pixellm} & 9.2 & 11.8 & 0.0 \\
PixelLM-13B~\cite{ren2024pixellm} & 12.8 & 15.4 & 0.0 \\
BiomedParse$^{\P}$~\cite{zhao2024biomedparse} & \underline{23.8} & 18.5 & 0.6 \\
RecLMIS$^{\P}$~\cite{huang2024cross} & 22.4 & \underline{19.5} & 0.0 \\
IMIS-Net$^{\P}$~\cite{cheng2025interactive} & 9.8 & 11.8 & \underline{21.6} \\
\midrule
\textbf{ROSALIA (Ours)} & \textbf{71.2} & \textbf{75.6} & \textbf{91.8} \\
\bottomrule
\end{tabular}
\end{adjustbox}
\label{tab:main result}
\vspace{-0.2cm}
\end{table}

%% file: table/main_lesion.tex
\begin{table}[h]
\vspace{-0.2cm}
\caption{Segmentation performance (\%) of ROSALIA for each lesion type.}
\vspace{-0.2cm}
\small
\centering
\begin{adjustbox}{max width=0.73\linewidth}
\begin{tabular}{c|ccc}
\toprule
Lesion & gIoU & cIoU & N-Acc. \\
\midrule
Cardiomegaly & 89.0 & 89.0 & 85.8 \\
Pneumonia & 57.2 & 60.4 & 97.1 \\
Atelectasis & 60.2 & 58.7 & 91.7 \\
Opacity & 60.5 & 64.2 & 85.0 \\
Consolidation & 61.9 & 65.6 & 91.2 \\
Edema & 64.8 & 66.6 & 92.2 \\
Effusion & 60.3 & 59.6 & 90.4 \\
\midrule
\textbf{Total} & \textbf{71.2} & \textbf{75.6} & \textbf{91.8} \\
\bottomrule
\end{tabular}
\end{adjustbox}
\label{tab:main result by lesion}
\vspace{0.1cm}
\end{table}

%% file: table/text_accuracy_class.tex
\begin{table}[h]
\vspace{-0.2cm}
\caption{Text response accuracy (\%) of ROSALIA.}
\vspace{-0.2cm}
\small
\centering
\begin{adjustbox}{max width=0.78\linewidth}
\begin{tabular}{c|c|ccc}
\toprule
Type & Overall & Basic & Global & Lesion Inf. \\
\midrule
Positive & 90.7 & 95.4 & 93.7 & 75.1 \\
Negative & 95.3 & 96.9 & 82.3 & 90.6 \\
\midrule
\textbf{Total} & \textbf{94.4} & \textbf{96.8} & \textbf{88.8} & \textbf{84.8} \\
\bottomrule
\end{tabular}
\end{adjustbox}
\label{tab:text_result}
\vspace{-0.2cm}
\end{table}

%% file: sec/5_conclusion.tex
\section{Conclusion}
\label{sec:conclusion}
In this study, we introduce \dataset, the first dataset for instruction-guided lesion segmentation in CXRs, along with \model, a VLM developed for this new task. Our automated pipeline enables the construction of this million-scale dataset, and expert evaluations show a remarkably high acceptance rate, confirming the quality and reliability of our fully human-free data generation process. Trained on \dataset, \model demonstrates a comprehensive ability to generate accurate lesion segmentations and textual responses across diverse user instructions. These findings indicate that \dataset and \model offer a strong foundation for advancing research on fine-grained lesion grounding in the CXR domain.

\clearpage

\paragraph{Acknowledgments.}
This work was supported by the Institute of Information \& Communications Technology Planning \& Evaluation (IITP) grants (No.RS-2019-II190075, No.RS-2022-II220984, No.RS-2024-00457882, No.RS-2025-02304967, No.RS-2025-02305581), the Korea Health Industry Development Institute (KHIDI) grant (No.RS-2025-02213750), and National Research Foundation of Korea (NRF) grant (NRF-2020H1D3A2A03100945, No.RS-2023-00209060), and the Medical Scientist Training Program funded by the Korean government (MSIT, MOHW).

%% file: sec/X_suppl.tex
\clearpage
\appendix   
\setcounter{page}{0} 
\addtocounter{linenumber}{212} 
\thispagestyle{empty}
\onecolumn           

\mtcsettitle{parttoc}{$<$Table of Contents$>$\\[15pt]}
{%
\renewcommand{\thepart}{} \renewcommand{\partname}{}  
\part{Appendix}       
}
\setbox0=\vbox{\tableofcontents}
\parttoc

\twocolumn           
\clearpage           

\section{Grounded Lesion Mask Generation}
\label{sec:suppl_report_grounding}

\subsection{Report Pre-Processing}
\label{sec:suppl_report_preprocessing}
Textual information for the CXR images from MIMIC-CXR was extracted from their corresponding radiology reports. From these raw reports, we extract the findings, impression, and last paragraph sections following the official MIMIC report pre-processing code. We then adhere to a hierarchical fallback logic to select a single representative text section for each study: the impression section is used if the findings section is missing, and the last paragraph is used if the impression is also absent. Studies that lack all three of these sections are excluded.

\subsection{Large Language Models and Prompts}
\label{sec:suppl_llms_and_prompts}
\begin{figure*}[t]
  \centering
  \includegraphics[width=0.8\linewidth]{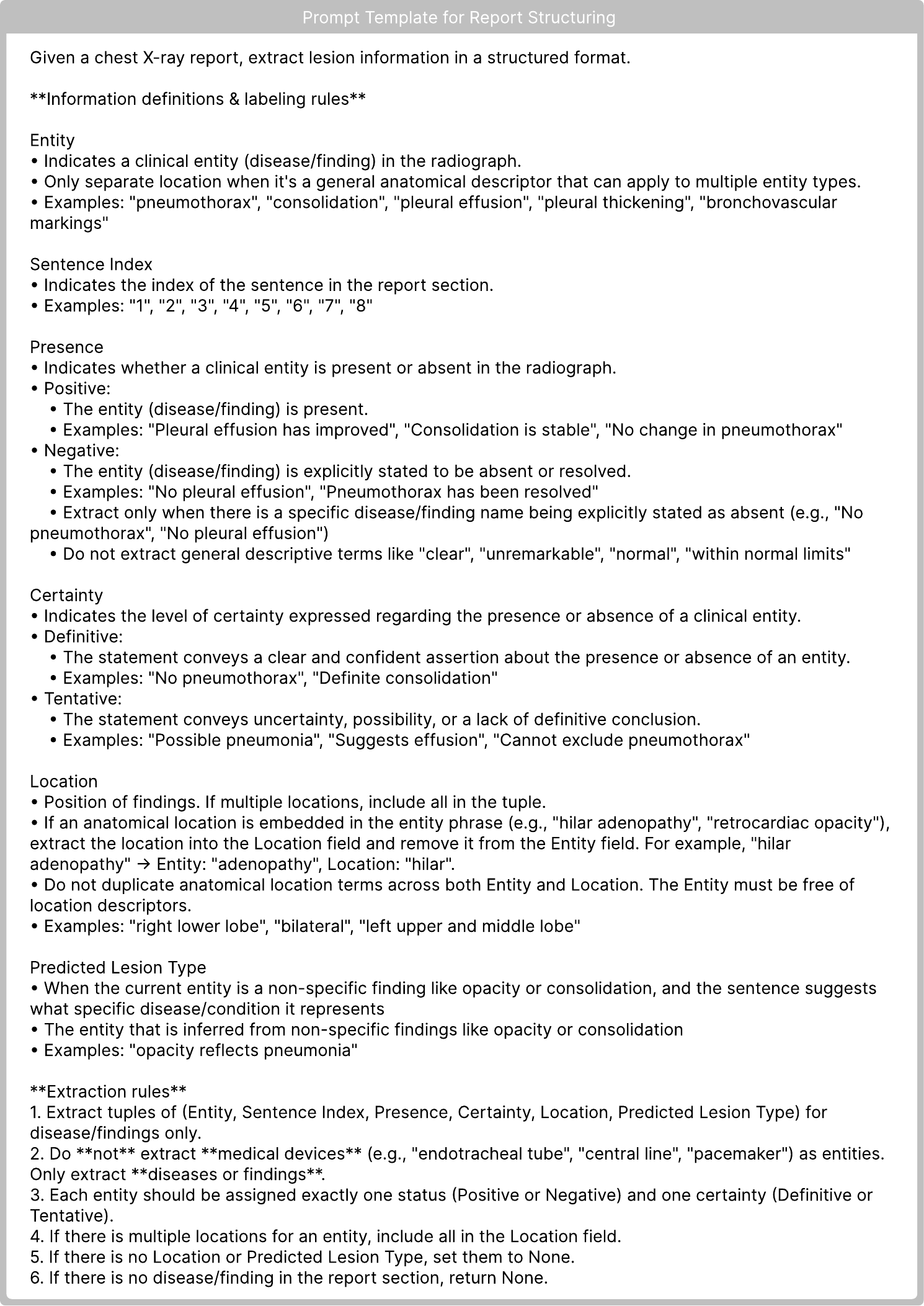}
  \vspace{0cm}
  \caption{A prompt template for report structuring.}
  \label{fig:instruction template}
  \vspace{0cm}
\end{figure*}

\begin{figure*}[t]
  \centering
  \includegraphics[width=0.8\linewidth]{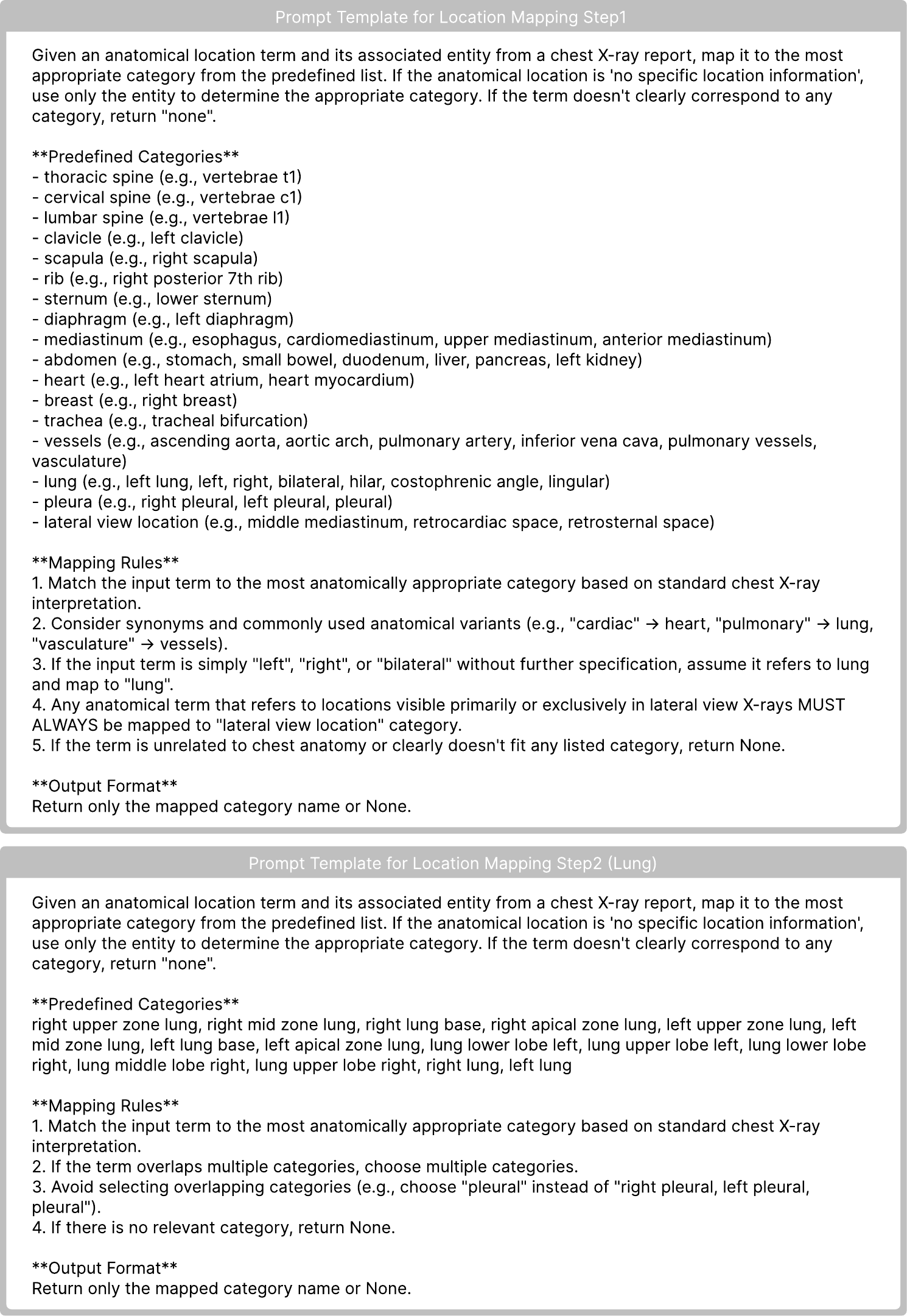}
  \vspace{0cm}
  \caption{A prompt template for location mapping. Step 2 illustrates the scenario following the `lung' mapping from Step 1.}
  \label{fig:instruction template2}
  \vspace{0cm}
\end{figure*}

To extract information from the pre-processed report section, our pipeline employs two distinct large language models (LLMs). The initial report structuring step utilizes Mistral-Small-3.1-24B-Instruct-2503. Using the prompt shown in Figure~\ref{fig:instruction template}, we extract lesion information from the report as six-element tuples. For the subsequent location mapping step, we employ medgemma-27b-text-it, which is specialized in the medical domain. Using the prompt shown in Figure~\ref{fig:instruction template2}, we normalize the lesion's location in a two-step process. In compliance with the PhysioNet credentialed data use agreement for MIMIC-CXR, both models were run on our local GPU setup. 

\subsection{Vision Models and Characteristics}
\label{sec:suppl_vision_models}
\vspace{0pt}
\noindent\textbf{Pretrained HybridGNet.} We utilized a HybridGNet model, pretrained on the CheXMask dataset~\cite{gaggion2024chexmask, gaggionchexmask, Gaggion_2022}, to segment the right lung, left lung, and heart. It demonstrates robust segmentation performance for these three organs, even in challenging CXRs from patients with severe conditions characterized by dense opacities.The resulting masks serve multiple, distinct roles in our pipeline. The heart mask is used directly as the ground-truth lesion mask for cardiomegaly. The right and left lung masks serve two purposes: they are used as the ($L_r, L_l$) inputs in Algorithm~\ref{alg:lesion_mask_generation}, and they are also merged with the heart mask to define the editing region for RadEdit. Details regarding the use of this model were omitted from the main text for brevity.

\vspace{10pt}
\noindent\textbf{RadEdit.} This diffusion-based image editing model takes a chest X-ray image and a text prompt as input~\cite{perez2024radedit}. To transform the input into a normal-appearing image, we used the standard prompt on which RadEdit was trained: ``No acute cardiopulmonary process''. Additionally, it requires a mask specifying the editing region. For this, we used the merged masks from the pretrained HybridGNet described above. Notably, we used the original MIMIC-CXR dataset~\cite{johnson2019mimic, johnson2024mimic}, which contains DICOM files, rather than the MIMIC-CXR-JPG version~\cite{johnson2019mimic_cxr_jpg}. This is because RadEdit was trained on the original MIMIC-CXR, and we observed that inputting the histogram-equalized MIMIC-CXR-JPG images significantly degraded the quality of the edited image.

\vspace{10pt}
\noindent\textbf{CXAS.} Designed for anatomy segmentation in CXRs, this model is capable of segmenting 159 anatomical region classes~\cite{seibold2023accurate}. Specifically, in our research, we input the opacity-removed images (the output of RadEdit) into CXAS to segment the anatomy. This is because CXAS tends to produce lower-quality anatomy masks for patients with significant opacities.

\vspace{10pt}
\noindent\textbf{Pretrained YOLO.} For lesion detection, we employed a YOLO model, specifically utilizing the checkpoint from the submitted solution in the VinBigData Chest X-ray Abnormalities Detection competition~\cite{vinbigdata-chest-xray-abnormalities-detection, nguyen2022vindr}. Although this model can detect various types of lesions (aortic enlargement, atelectasis, calcification, consolidation, ILD, infiltration, lung opacity, nodule/mass, other lesion, pleural effusion, pleural thickening, pneumothorax, pulmonary fibrosis), we filtered its outputs to retain only those findings considered hyperintense lesions. We therefore excluded aortic enlargement, other lesion, and pneumothorax from the detection categories.

\subsection{Thresholds for Lesion Mask Generation}
\label{sec:suppl_threshold}

The thresholds in Equation~\ref{eq:anomaly_set} and Algorithm~\ref{alg:lesion_mask_generation} were carefully calibrated to ensure maximum mask quality. The final values were determined through an iterative process involving multiple quality checks by a physician, who identified the settings that maximized the yield of high-quality masks. The finalized thresholds are summarized in Table~\ref{tab:thresholds}. With the exception of edema, the threshold settings are identical for all other lesion types. We set the threshold values for edema lower than for other lesion types because it tends to spread widely throughout the lungs.

\begin{table}[h!]
\centering
\caption{Threshold values used to generate the lesion masks. `General' lesions refer to all lesions other than edema.}
\label{tab:thresholds}
\begin{tabular}{lcc}
\toprule
\textbf{Threshold} & \textbf{General} & \textbf{Edema} \\
\midrule
$\tau_\mathrm{ano}$ & 0.10 & 0.01 \\
$\tau_\mathrm{anatomy}$ & 0.25 & 0.25 \\
$\tau_\mathrm{conf}$ & 0.20 & 0.01 \\
$\tau_\mathrm{signal}$ & 0.20 & 0.20 \\
$\tau_\mathrm{size}$ & 0.10 & 0.10 \\
\bottomrule
\end{tabular}
\end{table}

\begin{figure*}[t]
  \centering
  \includegraphics[width=0.98\textwidth]{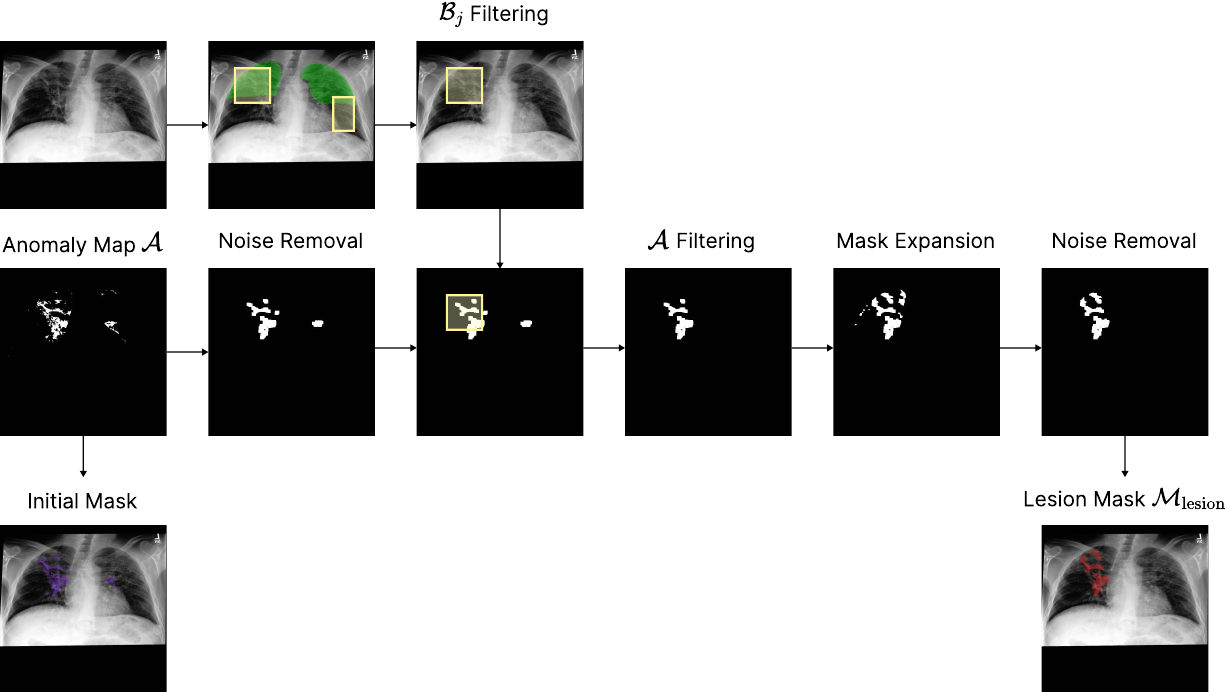}
  \caption{An example of detailed lesion mask generation where the report mentions ``Areas of streaky opacity are again seen in the upper lobes.'' but the mask is grounded only to the right upper lobe. In the top row, the yellow lesion box mask in the left lung is discarded due to insufficient overlap with the green upper lung mask. The remaining box mask in the right lung is then used to filter the anomaly map. Intermediate post-processing steps, including noise removal and mask expansion, are applied to enhance the final mask quality.}
  \label{fig:detailed_lesion_mask_generation}
  \vspace{-0.5cm}
\end{figure*}

\subsection{Lesion Mask Post-Processing}
\label{sec:suppl_post_processing}

To further enhance the quality of the final lesion masks, additional post-processing steps were applied. Sequential erosion and dilation operations are used to remove small, scattered noise. Although omitted from the main text for brevity, this noise removal step is also performed prior to filtering the anomaly map with the lesion box mask. We also expanded the lesion masks to include adjacent pixels with similar intensity values for more complete segmentation. Furthermore, specifically for effusions at the lung base, we incorporated the lower portion of the lung masks from the pretrained HybridGNet to ensure clean coverage extending to the costophrenic angle. The detailed process is illustrated in Figure~\ref{fig:detailed_lesion_mask_generation}.

\subsection{Empty Location}
\label{sec:suppl_empty_location}
We extract an ``empty location'' during the location verification step. An empty location is defined as a lung region where a specific lesion is not present. We designated a lung region as an empty location if it did not overlap at all with the anatomy masks corresponding to the reported location. To identify locations that are truly free of any reported lesions, we compute this not only for the seven major lesions but for all lesions mentioned in the report, and utilize this information in a subsequent data generation step. 

\subsection{Sample Discarding and Pipeline Recall}
\label{sec:suppl_pipeline_recall}
In our pipeline, samples that did not meet the strict cross-model consistency criteria were excluded, and the resulting recall rate is reported in Table~\ref{tab:recall}. While the recall can be flexibly increased by relaxing these criteria, our primary goal is to generate high-confidence samples rather than to maximize recall. Since training the LISA on a higher-recall dataset led to degraded performance on the MIMIC-ILS test set (gIoU: 54.3\%, cIoU: 61.4\%, N-Acc: 95.8\%), we opted for a high-threshold setting.

\vspace{-0.3cm}
\begin{table}[h!]
\centering
\caption{Recall comparison for different threshold values.}
\label{tab:recall}
\vspace{-0.3cm}
\resizebox{\columnwidth}{!}{
\begin{tabular}{lccccccc} 
\toprule
\textbf{Lesion} & \textbf{atel.} & \textbf{pneu.} & \textbf{effu.} & \textbf{opac.} & \textbf{edem.} & \textbf{cons.} & \textbf{Avg.} \\
\midrule
Recall (default $\tau$) & 13.4 & 28.5 & 16.0 & 21.6 & 58.0 & 31.9 & 28.3 \\
Recall (half $\tau$)    & 28.7 & 49.8 & 32.6 & 40.3 & 82.1 & 50.6 & 47.3 \\
\bottomrule
\end{tabular}%
}
\end{table}
\vspace{-0.3cm}

\section{Lesion Types}
\label{sec:suppl_lesion_types}
We construct our dataset around seven major lesion types commonly observed in CXRs, identified through discussions with board-certified physicians: opacity, consolidation, pnuemonia, atelectasis, edema, cardiomegaly, and effusion. These disease categories are widely utilized in many CXR-related studies~\cite{irvin2019chexpert, wang2017chestx, demner2015preparing}. First, we include \textit{opacity} and \textit{consolidation}, which are high-level, comprehensive terms referring to hyperintense lesions. These broad categories can be mapped to specific lung lesion types, including \textit{pneumonia}, \textit{atelectasis}, and \textit{edema}. We also include two major non-lung disease categories: \textit{cardiomegaly}, the enlargement of the heart, and \textit{(pleural) effusion}, the accumulation of fluid in the pleural space. 

\clearpage

\clearpage
\vspace{-0.5cm}
\section{Instruction-Answer Pair Generation}
\label{sec:suppl_instruction_answer_pair_generation}

\subsection{Positive Instruction}
\label{sec:suppl_positive_instruction}

\noindent\textbf{Basic Instruction.} The instructions for positive samples are generated directly from the grounded lesion mask generation results. For instance, if a definitive finding of pneumonia has a grounded location of right lung base and left lung base, we generate a basic instruction: ``Segment the pneumonia in the right lung base and left lung base.''.
However, if the finding's certainty is tentative, indicating the lesion's presence is not definitive, we substitute it with the more general term opacity to create the basic instruction. For example, the previous instruction becomes: ``Segment the opacity in the right lung base and left lung base.''. As listed in Table~\ref{tab:location_category}, the target location can be specified as a single area—either a broad region or a specific lung zone—or as a combination of these areas.

\input{table/location_category}

\vspace{10pt}
\noindent\textbf{Global Instruction.} A global instruction is used to segment all instances of a lesion across the entire lung, without specifying a location. To create a valid global instruction, the generated lesion must cover all lesions cited in the report; this ensures the mask can serve as a complete ground truth. Therefore, we only generate global instructions when the \textit{grounded location}, where masks were actually generated, and the \textit{reported location}, the complete area mentioned in the report, are identical. If the lesion type is cardiomegaly, we always generate this instruction type. This is because cardiomegaly represents a condition of the heart itself, rather than a lesion that can appear in variable locations.

\vspace{10pt}
\noindent\textbf{Lesion Inference Instruction.}
We generate lesion inference instructions to enable the model to infer the specific lesion type from an opacity at a given location. These instructions are generated for findings regardless of their original certainty level, as the certainty is instead reflected in the ground-truth text description. We selected pneumonia, atelectasis, and edema as the target lesion types for this task. This choice reflects clinical reporting practices, where radiologists often describe these specific findings using an inferential process. In contrast, other major lesion types are typically stated directly. For example, a report rarely states, ``There is an opacity in the left lung. It is highly suggestive of effusion.''; instead, the finding is stated directly as ``Left lung effusion.''

\subsection{Negative Instruction}
\label{sec:suppl_negative_instruction}
Negative instructions are generated in two main scenarios. First, we generate instructions for lesions that are either never mentioned or negated (\eg, ``no pneumonia'') in the report. For these findings, we create a negative instruction, which can be either a basic type by randomly assigning a lung region (\eg, ``Segment the pneumonia in the left lung.''), or a global type (\eg, ``Segment the pneumonia''). The second method involves pairing a target lesion with a randomly selected empty location. For example, if right lung apex and left lung apex are empty locations, we can generate the instruction, ``Segment the atelectasis in the right lung apex.'' To prevent an excessive number of negative samples, our logic restricts the generation to a maximum of one negative instruction per lesion type for each study.

\subsection{Clinical Utility of MIMIC-ILS}
\label{sec:suppl_clinical_utility}
\noindent\textbf{Diversity of Instructions.} The functional scope of the dataset is driven primarily by the diversity of disease–anatomy combinations rather than by the number of instruction templates. Linguistic diversity can be readily addressed by paraphrasing existing instructions in MIMIC-ILS. To demonstrate this feasibility, we used Qwen3-Next-80B-A3B-Instruct to paraphrase each original instruction into nine variants reflecting three user personas (medical experts, laypersons, and AI developers). LISA trained on this enriched dataset still demonstrate strong performance (gIoU: 67.3\%, cIoU: 73.1\%, N-Acc: 96.5\%) on the paraphrased test set.

\vspace{10pt}
\noindent\textbf{Usability for Laypersons.} Users without medical expertise cannot be expected to visually identify lesions in a scan to provide basic instructions. However, because MIMIC-ILS incorporates negative cases for absence confirmation, a model trained on this dataset naturally overcomes this limitation. By iteratively querying the model across various anatomical locations, the system can autonomously verify the presence of a lesion—outputting a precise segmentation mask if it exists, or confirming its absence otherwise. Similarly, global instructions (e.g., ``Segment the opacity") offer an intuitive way for users to make broad inquiries. Coupled with the model's robust handling of negative cases, these capabilities ensure that users can effectively obtain screening results without ever needing to visually inspect the image themselves.

\clearpage
\section{Quality Control}
\label{sec:mimic_quality_control}

\subsection{Chest X-rays}
\label{sec:chest_xrays}
We exclusively utilized Posteroanterior (PA) and Anteroposterior (AP) view images from the MIMIC-CXR dataset. However, even within these designated views, the dataset contains noisy samples, including mislabeled lateral views, non-chest X-rays, or images with severe anatomical truncation. To ensure data quality, we leveraged metadata from CXReasonBench~\cite{lee2025cxreasonbench, lee2025cxreasonbenchdataset}. This dataset was meticulously constructed from frontal view images within the MIMIC-CXR dataset that had verified high image quality. Specifically, we utilized its pre-extracted information such as the count of extractable CXAS anatomy masks and indicators of full chest visibility to identify and exclude these problematic images beforehand.

\subsection{Lung and Heart Masks}
\label{sec:qc_lung_and_heart_masks}
Lung and heart masks are a critical component for the construction of MIMIC-ILS. However, both models can produce erroneous results: the pretrained HybridGNet occasionally generates abnormal masks, and CXAS (even when applied to RadEdit-processed images) also generates suboptimal masks. To address this, we cross-referenced the masks from both models and excluded cases with significant discrepancies, interpreting this as a failure in either the HybridGNet or CXAS segmentation. Specifically, we determined that large differences in the outermost x-coordinates of the lung masks or the lowermost y-coordinates of the heart masks would cause problems for subsequent grounded lesion mask generation, and thus excluded these studies.

\subsection{Cardiomegaly}
\label{sec:qc_cardiomegaly}
To generate reliable negative samples for cardiomegaly, we measured the cardiothoracic ratio (CTR) using the right lung, left lung, and heart masks generated by the pretrained HybridGNet. We then filtered these samples, exclusively including those with a CTR of 0.45 or less in our final negative dataset. This 0.45 threshold was calibrated by a physician who analyzed the distribution of CXRs across different CTR intervals to establish a clinically sound cutoff.

\newpage
\section{MIMIC-ILS Dataset}
\label{sec:mimic_ils_dataset}
\subsection{Details for Dataset Splits} 
\label{sec:dataset split detail} 
The data splits and distribution by instruction type for MIMIC-ILS are presented in Tables~\ref{tab:qa_statistics_train}, \ref{tab:qa_statistics_val}, and \ref{tab:qa_statistics_test}. Note that the counts for the test set reflect the final numbers after excluding cases that were rejected during the quality assessment process.

\subsection{Quality Assessment} \label{sec:quality_assessment} A rigorous quality assessment was conducted on the test split by four physicians. All positive samples were reviewed by all four physicians, while the negative samples were divided among them for evaluation. The reviewers were provided with an CXR image, the lesion type, and the mapped anatomical location text generated from our information-grounding process, along with the corresponding ground-truth radiology report. They were then asked to mark each pair as either “Acceptable” or “Not Acceptable” on a review sheet. The evaluation process was conducted independently for each expert, ensuring that no reviewer could access the others’ evaluation results.

\subsection{Details on Expert Evaluation} \label{sec:medical expert profiles}
The expert evaluations were conducted by four physicians, all experienced radiation oncologists with extensive training in lesion contouring. Their professional backgrounds are as follows: Experts A and B are board-certified physicians with 9 and 7 years of clinical experience, respectively, while Experts C and D are resident doctors, each with 6 years of clinical experience. Also, the lesion-level acceptance rate in human evaluation are shown in Table~\ref{tab:human_eval_lesion}. 

\input{table/mimic_ils_train}
\input{table/mimic_ils_val}
\input{table/mimic_ils_test}
\input{table/human_eval_lesion}

\subsection{Data Generation Examples} \label{sec:examples}
\begin{figure*}[t]
  \centering
  \includegraphics[width=0.99\linewidth]{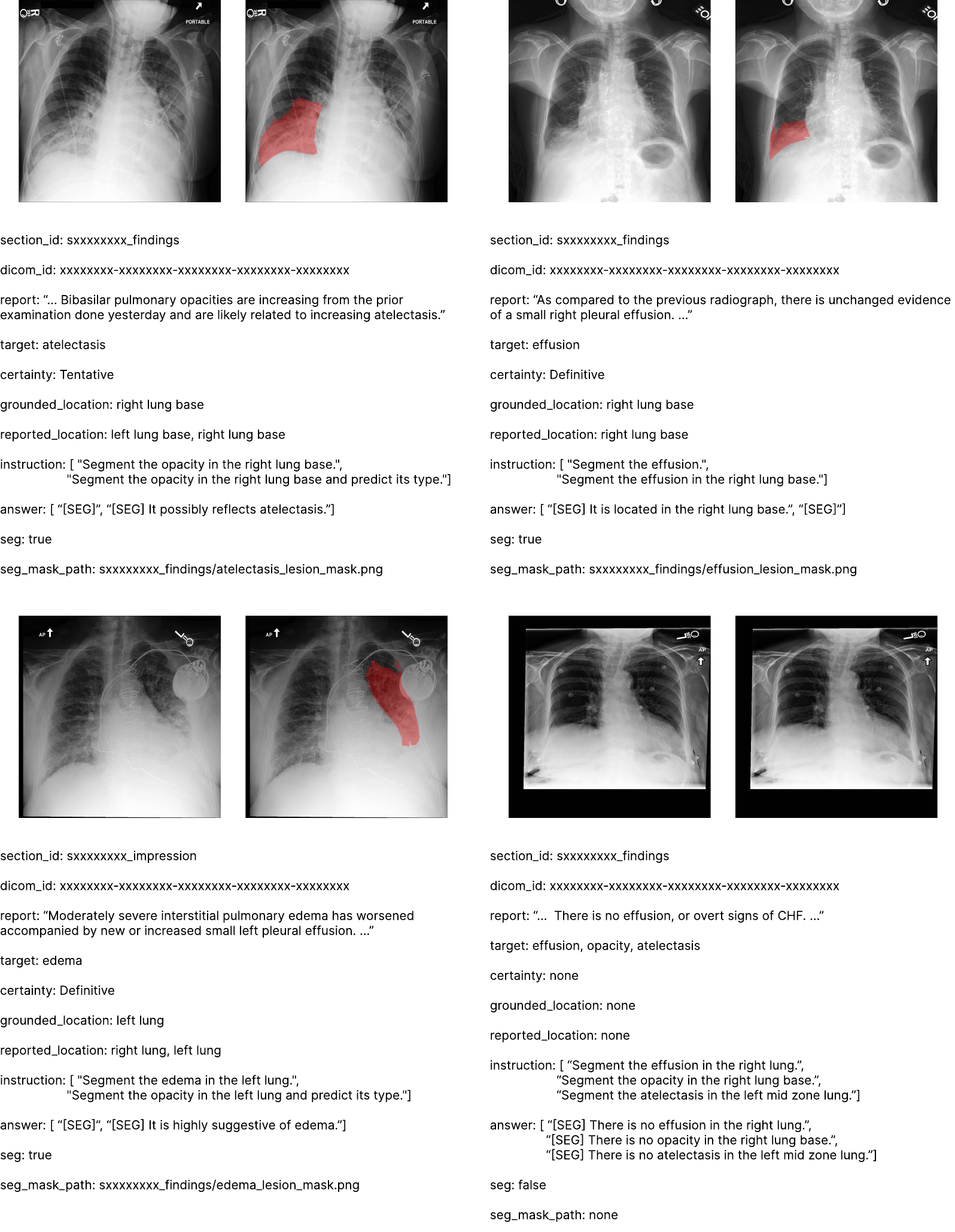}
  \vspace{0cm}
  \caption{Examples of final generated samples in our MIMIC-ILS dataset.}
  \label{fig:mimic_ils_example_supple}
  \vspace{0cm}
\end{figure*}

With our proposed data generation pipeline, we can produce high-quality lesion masks and their corresponding instruction–answer pairs from grounded information. Figure~\ref{fig:mimic_ils_example_supple} illustrates representative examples across various lesion types and anatomical locations, including both positive and negative cases, along with their corresponding structured information.

\clearpage
\section{Model Training Details}
\label{sec:suppl_model_training}

In our experiments, training the LISA-7B–based model took approximately two and a half days on two NVIDIA H100 GPUs for 15 epochs. Using the DeepSpeed package~\cite{rasley2020deepspeed}, we trained the model with the DeepSpeed Stage-2 configuration and a WarmupDecayLR scheduler, with 100 warmup steps and a minimum and maximum learning rate of 0 and 0.0003, respectively. For inference on the test set, which contains 12K examples, segmentation alone takes about 20 minutes, whereas segmentation with text outputs requires approximately 1.5 hours. During training, each input image had a 50\% chance of being processed with histogram equalization.

\newpage
\section{Additional Experimental Results}
\label{sec:additional result}
\subsection{Lesion-Wise Text Accuracy} 
\label{sec:lesion-wise result}
The text-response accuracy of ROSALIA for each lesion type is summarized in Table~\ref{tab:text_result_lesion}. Across most lesion and question types, the model consistently achieves high accuracy, similar to the segmentation performance reported in Table~\ref{tab:main result by lesion}. For lesion-inference questions, CXR alone typically cannot provide a definitive diagnosis and often requires additional examinations (e.g., blood tests or cultures). As a result, radiologists generally provide only a differential diagnosis based on visual findings. Given this inherent uncertainty in CXR interpretation, improvements in accuracy for lesion-inference questions are naturally limited. Nevertheless, we evaluated how well the trained model on our dataset can perform on this question type and leave further advancements in this direction to future work.

\input{table/text_accuracy_lesion}

\subsection{Additional Qualitative Examples} 
\begin{figure*}[t]
  \centering
  \includegraphics[width=0.99\linewidth]{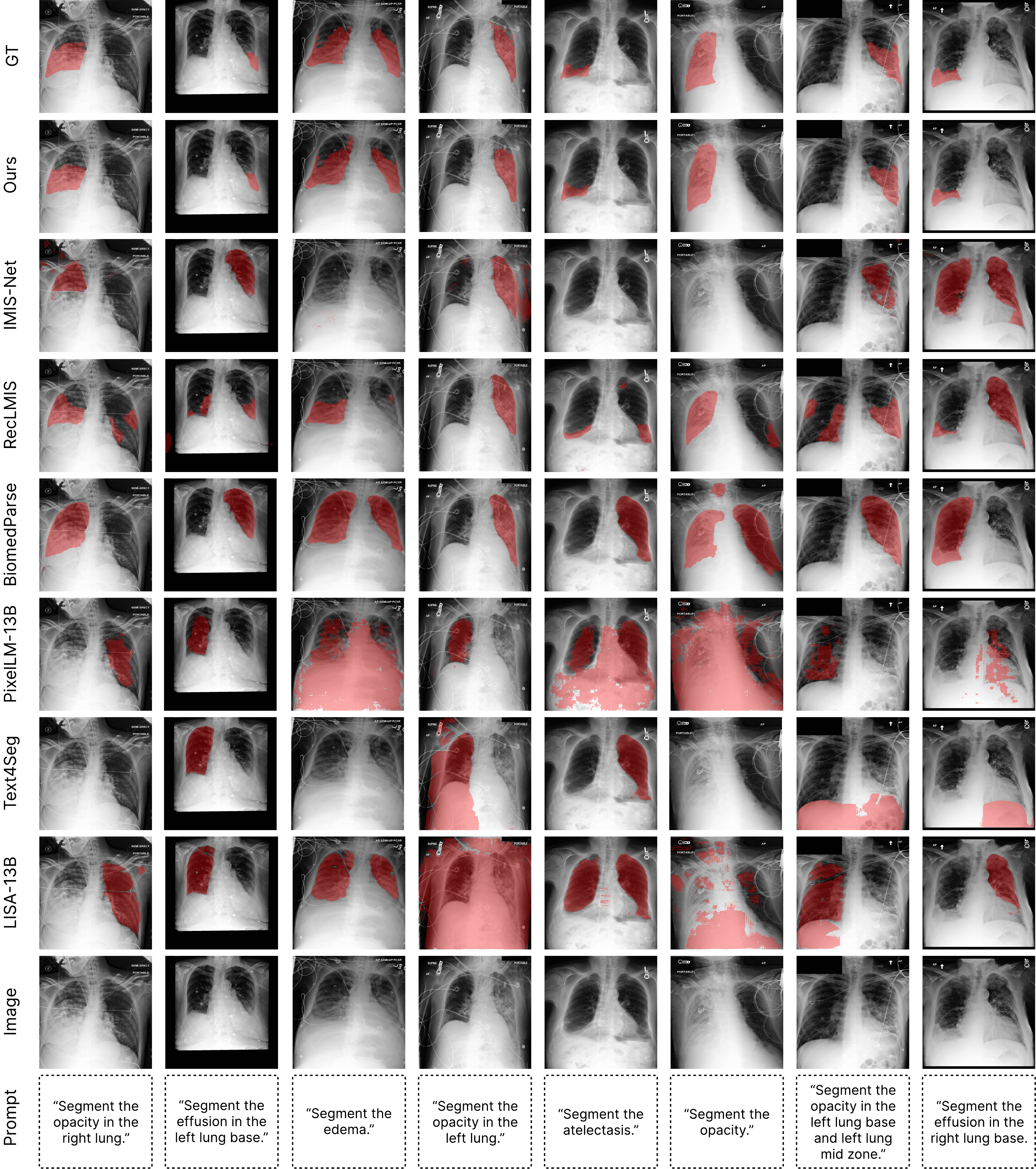}
  \vspace{0cm}
  \caption{Qualitative comparison of segmentation results between ROSALIA and baseline models.}
  \label{fig:grid1_supple}
  \vspace{0cm}
\end{figure*}

\begin{figure*}[t]
  \centering
  \includegraphics[width=0.99\linewidth]{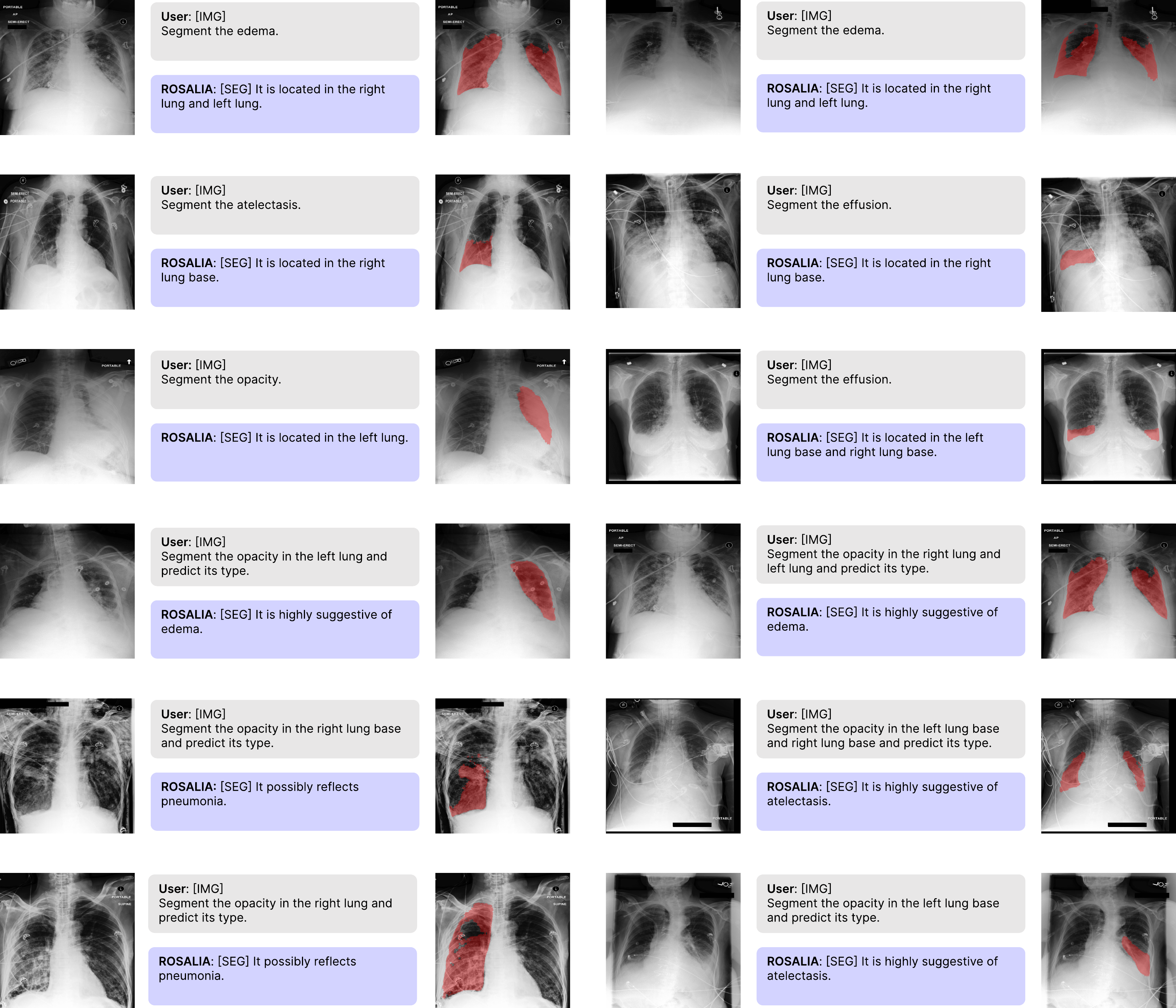}
  \vspace{0cm}
  \caption{Examples of textual responses generated by ROSALIA. All generated text responses correctly match the ground-truth answers, and both the segmentation and textual outputs in this figure are rated as good examples by medical experts.}
  \label{fig:text_response_supple}
  \vspace{0cm}
\end{figure*}

We present additional qualitative examples comparing ROSALIA with baseline models in Figure~\ref{fig:grid1_supple}. Unlike the baselines, ROSALIA produces accurate segmentation outputs tailored to diverse user instructions. In addition, examples that include both text responses and segmentation outputs are shown in Figure~\ref{fig:text_response_supple}. In these cases as well, ROSALIA provides highly factual text responses alongside precise segmentation results.

\clearpage

\clearpage

%% file: table/location_category.tex
\begin{table}[h]
\centering
\caption{List of valid target locations for basic instructions. Locations are categorized into broad regions and specific lung zones.}
\label{tab:location_category}
\begin{adjustbox}{max width=0.75\linewidth}
\begin{tabular}{l|l}
\toprule
\textbf{Lung Region} & \textbf{Location Name} \\
\midrule
\multirow{2}{*}{\textbf{Broad Regions}} & right lung \\
& left lung \\
\midrule
\multirow{4}{*}{\textbf{Lung Zones (Right)}} & right apical zone lung\\
& right upper zone lung\\
& right mid zone lung\\
& right lung base \\
\midrule
\multirow{4}{*}{\textbf{Lung Zones (Left)}} & left apical zone lung\\
& left upper zone lung\\
& left mid zone lung\\
& left lung base \\
\bottomrule
\end{tabular}
\end{adjustbox}
\end{table}

%% file: table/mimic_ils_train.tex
\begin{table*}[t]
\caption{Number of generated instruction-answer pairs per lesion and template type in MIMIC-ILS train split.}
\vspace{-0.2cm}
\small
\centering
\begin{adjustbox}{max width=\textwidth}
\begin{tabular}{c|c|cc|cc|cc} 
\toprule
\multirow{2}{*}{Lesion} & \multirow{2}{*}{\# IAs} & \multicolumn{2}{c|}{Basic} & \multicolumn{2}{c|}{Global} & \multicolumn{2}{c}{Lesion Inference} \\
\cmidrule(lr){3-4} \cmidrule(lr){5-6} \cmidrule(lr){7-8}
 &  & pos & neg & pos & neg & pos & neg \\ 
\midrule
cardiomegaly  & 63,153  & 0     & 0      & 39,108 & 24,045 & 0     & 0 \\
pneumonia     & 158,059 & 4,542  & 145,317 & 511   & 3,147  & 4,542  & 0 \\
atelectasis   & 166,935 & 8,846  & 128,943 & 3,331  & 16,969 & 8,846  & 0 \\
opacity       & 156,807 & 9,113  & 73,619 & 1,532  & 274   & 0     & 72,269 \\
consolidation & 154,955 & 3,428  & 144,489 & 379   & 6,659  & 0     & 0 \\
edema         & 182,233 & 14,150 & 145,251 & 5,390  & 3,292  & 14,150 & 0 \\
effusion      & 162,998 & 10,244 & 114,375 & 3,713  & 34,666 & 0     & 0 \\
\midrule
Total         & 1,045,140 & 50,323 & 751,994 & 53,964 & 89,052 & 27,538 & 72,269 \\
\bottomrule
\end{tabular}
\end{adjustbox}
\label{tab:qa_statistics_train}
\vspace{0cm}
\end{table*}

%% file: table/mimic_ils_val.tex
\begin{table*}[t]
\vspace{-0.5cm}
\caption{Number of generated instruction-answer pairs per lesion and template type in MIMIC-ILS validation split.}
\vspace{-0.2cm}
\small
\centering
\begin{adjustbox}{max width=\textwidth}
\begin{tabular}{c|c|cc|cc|cc} 
\toprule
\multirow{2}{*}{Lesion} & \multirow{2}{*}{\# IAs} & \multicolumn{2}{c|}{Basic} & \multicolumn{2}{c|}{Global} & \multicolumn{2}{c}{Lesion Inference} \\
\cmidrule(lr){3-4} \cmidrule(lr){5-6} \cmidrule(lr){7-8}
 &  & pos & neg & pos & neg & pos & neg \\ 
\midrule
cardiomegaly  & 539   & 0   & 0    & 332 & 207 & 0   & 0 \\
pneumonia     & 1,211  & 28  & 1,130 & 2   & 23  & 28  & 0 \\
atelectasis   & 1,316  & 75  & 986 & 33  & 147 & 75  & 0 \\
opacity       & 1,225  & 75  & 581 & 13  & 0   & 0   & 556 \\
consolidation & 1,213  & 30  & 1,137 & 3   & 43  & 0   & 0 \\
edema         & 1,469  & 129 & 1,124 & 59  & 28  & 129 & 0 \\
effusion      & 1,273  & 416  & 885  & 22  & 287 & 0   & 0 \\
\midrule
Total         & 8,246  & 416 & 5,843 & 464 & 735 & 232 & 556 \\
\bottomrule
\end{tabular}
\end{adjustbox}
\label{tab:qa_statistics_val}
\vspace{0cm}
\end{table*}

%% file: table/mimic_ils_test.tex
\begin{table*}[t]
\vspace{-0.5cm}
\caption{Number of generated instruction-answer pairs per lesion and template type in MIMIC-ILS test split.}
\vspace{-0.2cm}
\small
\centering
\begin{adjustbox}{max width=\textwidth}
\begin{tabular}{c|c|cc|cc|cc} 
\toprule
\multirow{2}{*}{Lesion} & \multirow{2}{*}{\# IAs} & \multicolumn{2}{c|}{Basic} & \multicolumn{2}{c|}{Global} & \multicolumn{2}{c}{Lesion Inference} \\
\cmidrule(lr){3-4} \cmidrule(lr){5-6} \cmidrule(lr){7-8}
 &  & pos & neg & pos & neg & pos & neg \\
\midrule
cardiomegaly  & 965   & 0   & 0     & 803 & 162 & 0   & 0 \\
pneumonia     & 1,767 & 60  & 1,596 & 8  & 43  & 60  & 0 \\
atelectasis   & 1,842 & 110 & 1,466 & 45  & 111 & 110 & 0 \\
opacity       & 1,753 & 174 & 779 & 26  & 5   & 0   & 769 \\
consolidation & 1,756 & 69  & 1,612 & 10  & 65  & 0   & 0 \\
edema         & 2,274 & 283 & 1,551 & 103 & 54  & 283 & 0 \\
effusion      & 1,878 & 156 & 1,312 & 54  & 356 & 0   & 0 \\
\midrule 
Total         & 12,235 & 852 & 8,316 & 1,049 & 796 & 453 & 769 \\
\bottomrule
\end{tabular}
\end{adjustbox}
\label{tab:qa_statistics_test}
\vspace{0cm}
\end{table*}

%% file: table/human_eval_lesion.tex
\begin{table*}[h]
\vspace{-0.5cm}
\caption{Acceptance rate (\%) by lesion type for each expert in the human evaluation.}
\vspace{-0.2cm}
\small
\centering
\setlength{\tabcolsep}{5pt}
\begin{adjustbox}{max width=\textwidth}
\begin{tabular}{l|cc|c|cc|c|cc|c|cc|c}
\toprule
\multirow{2}{*}{Lesion} & 
\multicolumn{3}{c|}{Expert A} & 
\multicolumn{3}{c|}{Expert B} & 
\multicolumn{3}{c|}{Expert C} &  
\multicolumn{3}{c}{Expert D} \\
\cmidrule(lr){2-4} \cmidrule(lr){5-7} \cmidrule(lr){8-10} \cmidrule(lr){11-13}
 & Pos & Neg & Total 
 & Pos & Neg & Total 
 & Pos & Neg & Total 
 & Pos & Neg & Total \\
\midrule
Cardiomegaly 
& 97.7 & 97.1 & 97.7 
& 99.2 & 100.0 & 99.2 
& 100.0 & 100.0 & 100.0 
& 99.4 & 100.0 & 99.4 \\
Pneumonia & 90.0 & 98.5 & 97.3 
& 97.1 & 99.7 & 99.4 
& 98.6 & 99.5 & 99.4 
& 98.6 & 98.8 & 98.7 \\
Atelectasis & 97.2 & 98.8 & 98.4 
& 79.9 & 99.5 & 94.2 
& 100.0 & 99.0 & 99.3 
& 99.3 & 97.3 & 97.8 \\
Opacity & 92.6 & 92.3 & 92.4 
& 96.5 & 96.5 & 95.1 
& 99.5 & 92.7 & 96.0 
& 96.0 & 96.6 & 96.3 \\
Consolidation & 97.2 & 95.7 & 95.9 
& 100.0 & 97.2 & 97.6 
& 100.0 & 98.3 & 98.5 
& 100.0 & 99.0 & 99.2 \\
Edema & 95.8 & 95.1 & 95.4 
& 97.9 & 100.0 & 99.0 
& 100.0 & 97.3 & 98.5 
& 89.8 & 98.2 & 94.4 \\
Effusion & 89.8 & 96.6 & 94.1 
& 89.3 & 97.3 & 94.3 
& 99.5 & 97.6 & 98.3 
& 95.4 & 98.8 & 97.6 \\
\bottomrule
\end{tabular}
\end{adjustbox}
\label{tab:human_eval_lesion}
\vspace{-0.5cm}
\end{table*}

%% file: table/text_accuracy_lesion.tex
\begin{table}[h]
\caption{Text response accuracy (\%) of ROSALIA across different question and lesion types.}
\vspace{-0.2cm}
\small
\centering
\begin{adjustbox}{max width=0.95\linewidth}
\begin{tabular}{c|c|ccc}
\toprule
Lesion & Overall & Basic & Global & Lesion Inf. \\
\midrule
Cardiomegaly & 96.0 & - & 96.0 & - \\
Pneumonia & 96.3 & 99.2 & 72.6 & 36.7 \\
Atelectasis & 92.9 & 96.9 & 69.2 & 69.1 \\
Opacity & 91.5 & 92.2 & 93.6 & 90.5 \\
Consolidation & 97.3 & 97.6 & 92.0 & - \\
Edema & 93.4 & 96.2 & 74.5 & 85.5 \\
Effusion & 94.5 & 96.9 & 85.9 & - \\
\midrule
\textbf{Total} & \textbf{94.4} & \textbf{96.8} & \textbf{88.8} & \textbf{84.8} \\
\bottomrule
\end{tabular}
\end{adjustbox}
\label{tab:text_result_lesion}
\vspace{0cm}
\end{table}